\pdfoutput=1

\documentclass[10pt,letterpaper]{article}
\usepackage[top=0.85in,left=2.75in,footskip=0.75in]{geometry}

\usepackage{amsmath,amssymb}

\usepackage{changepage}

\usepackage{textcomp,marvosym}

\usepackage{cite}
\usepackage{multirow}
\usepackage{nameref,hyperref}

\usepackage{subcaption}

\usepackage[right]{lineno}

\usepackage[nopatch=eqnum]{microtype}
\DisableLigatures[f]{encoding = *, family = * }

\usepackage[table]{xcolor}

\usepackage{array}
\usepackage{algorithm}
\usepackage{algpseudocode}

\newcolumntype{+}{!{\vrule width 2pt}}

\newlength\savedwidth

\raggedright
\usepackage[aboveskip=1pt,labelfont=bf,labelsep=period,justification=raggedright,singlelinecheck=off]{caption}
\usepackage[normalem]{ulem}

\makeatletter
\renewcommand{\@biblabel}[1]{\quad#1.}
\makeatother

\usepackage{lastpage,fancyhdr,graphicx}
\usepackage{epstopdf}
\fancyheadoffset[L]{2.25in}
\fancyfootoffset[L]{2.25in}
\newcommand{\bluetext}[1]{\textcolor{blue}{#1}}

\begin{document}
\vspace*{0.2in}

\begin{flushleft}
{\Large
\textbf\newline{Path Planning on Multi-level Point Cloud with a Weighted Traversability Graph} 
}
\newline
\\
Yujie Tang\textsuperscript{3},
Quan Li\textsuperscript{1},
Hao Geng\textsuperscript{1},
Yangmin Xie\textsuperscript{1,2},
Yusheng Yang\textsuperscript{1,2}
Hang Shi\textsuperscript{1*,4},
\\
\bigskip
\textbf{1} School of Mechatronics Engineering and Automation, Shanghai University, Shanghai, 200444, China
\\
\textbf{2} Shanghai Key Laboratory of Intelligent Manufacturing and Robotics, Shanghai University, Shanghai, 200444, China
\\
\textbf{3} Areaspace Information Research Institute, Chinese Academy of Sciences, Beijing, China
\\
\textbf{4} Jiangxi Research Institute, Beihang University, Nanchang 330200, China
\\
\bigskip

%
%





* hangshi@shu.edu.cn

\end{flushleft}
\section*{Abstract}
This article proposes a new path planning method for addressing multi-level terrain situations. The proposed method includes innovations in three aspects: 1) the pre-processing of point cloud maps with a multi-level skip-list structure and data-slimming algorithm for well-organized and simplified map formalization and management, 2) the direct acquisition of local traversability indexes through vehicle and point cloud interaction analysis, which saves work in surface fitting, and 3) the assignment of traversability indexes on a multi-level connectivity graph to generate a weighted traversability graph for generally search-based path planning. Then a modified A* algorithm is employed to exploit this traversability graph and generate short, safe paths across level transitions such as ramps, bridges, and overpasses. The proposed map structure reduces memory usage by over 70\% and achieves a 2× speedup in surface retrieval compared to state-of-the-art approaches designed for multi-layered environments. The planner attains a 100\% success rate across all evaluated scenarios, including challenging multi-level environments where conventional single-height 2.5D planners fail. The effectiveness and reliability of the proposed method are verified through indoor and outdoor experiments conducted in various environments, including multi-floor buildings, woodland, and rugged mountainous regions. The results demonstrate that the proposed method can properly address 3D path planning problems for ground vehicles in a wide range of situations.

\section{Introduction}\label{sec:Introduction}

Path planning in complex terrains remains a captivating research topic for mobile robots. Existing path planning techniques have primarily focused on dealing with occupancy maps in well-structured indoor environments or road driving scenarios, which have been extensively studied and implemented \cite{marder2010office,colas20133d}. However, terrains characterized by uneven and rough surfaces require explicit traversability assessment \cite{brunner2013rough,papadakis2013terrain,julia2012comparison}, typically encoded through traversability or cost maps \cite{borgessurvey2022}. Since ground robots operate primarily on terrain surfaces with movement constrained to SE(2) space, these traversability maps are commonly represented as 2D grid maps or 2.5D elevation maps \cite{xu2023efficient, feng2025constraint, qiu2024real}. However, when multi-layered structures arise in unstructured environments—such as bridges, tunnels, overpasses, or other vertically overlapping surfaces—conventional 2D representations encounter fundamental limitations due to height ambiguities at identical horizontal coordinates \cite{liu2023hybrid}.

Two primary approaches have been explored for path planning in multi-level terrains. The first approach involves the utilization of multi-level surface (MLS) maps, initially proposed by \cite{triebel2006multi}. MLS maps are 2.5D maps that incorporate level labels on vertical cells to represent the environmental structure or locate vehicles on multi-layered terrains. A notable study showcasing the use of MLS in path planning is presented in \cite{kummerle2009autonomous}, where an MLS map is constructed to separate the current level surface from the entire space. However, this approach primarily focuses on level segmentation, simplifying the problem as 2D local path planning after extracting the map of the current level. Consequently, it does not inherently address the challenge of multi-level path planning, where scenarios arise, for instance, when the vehicle and target are situated on different levels. 

The second approach performs path search directly in 3D space using nodes defined as 3D voxels \cite{colas20133d,menna2014real} or surface-projected poses \cite{krusi2017driving}, enabling level-agnostic 3D path planning. To mitigate the computational burden of increased dimensionality, these methods typically restrict planning to relevant locations—the terrain surfaces themselves. For instance, \cite{krusi2017driving} projects 6-DOF robot poses onto terrain surfaces and executes sampling-based planning directly on raw point clouds. While straightforward and memory-efficient in representation, integrating terrain assessment into planning significantly increases computation time due to the volumetric point cloud data, particularly in large-scale environments. \cite{atas2022elevation} addresses this through surfel representation of point clouds, forming a surfel-based state space for planning with sampling strategies. Nevertheless, this approach requires expensive preprocessing (RANSAC plane fitting and k-NN graph construction) and achieves lower success rates depending on sampling configuration. \cite{wang2023towards} adopts a 3D occupancy map with a Valid Ground Filter to extract traversable surfaces, combined with a downward projection strategy that maps robot poses to the nearest surface below. However, this downward projection inherently assumes single-valued height functions $z=f(x,y)$, preventing the planner from correctly handling scenarios with vertical occlusion such as bridges, overpasses, and tunnels, where multiple traversable surfaces exist at the same horizontal position.

Targeting practical path planning for multi-level unstructured terrains, we propose an efficient multi-level 2.5D representation that achieves the representational completeness of 3D methods while preserving the computational efficiency of 2D approaches. Rather than forcing a choice between 2D efficiency and 3D completeness, we extend the 2D grid structure to store multiple vertical surface hypotheses per cell. Each cell maintains a compact list of layers, where each layer encodes height and traversability cost derived from local terrain geometry. Planning is formulated as graph search over a 2D lattice augmented with layer indices — maintaining 
$O(xy)$ base complexity while supporting seamless level transitions when geometric constraints permit. Our core contributions are:

(1) A multi-level 2.5D map structure with O(xy·L) complexity that reduces memory usage by over 70\% and achieves more than a 2× speedup in surface retrieval compared to surfel- and voxel-based methods~\cite{atas2022elevation,wang2023towards}, while explicitly modeling bridges, tunnels, and other multi-layer structures beyond the capability of single-height 2.5D maps.

(2) An integrated graph-based planning framework where terrain geometry and traversability are unified in a single representation, enabling standard A* search with automatic level transition reasoning. The planner achieves a 100\% success rate across all evaluated scenarios, including challenging multi-level structures such as bridges and multi-story parking garages, whereas a representative 2.5D planner (OMDP, \cite{wang2023towards}) fails in 1 out of 3 scenarios due to height ambiguity, and a state-of-the-art 3D surfel-based planner (SBSP, \cite{atas2022elevation}) fails when processing large-scale maps.

(3) Comprehensive validation across various terrains (including indoor, highway, woodland, and rough valley) and diverse multi-layered scenarios (highway overpasses and parking structures), demonstrating robustness to various terrain surfaces and structural configurations.

The article is structured as follows. Section 2 presents the construction of the ML-SkiMap, while Section 3 introduces the traversability analysis. In Section 4, a modified version of the A* algorithm is presented, which is used to find feasible paths on the point cloud maps. In Section 5, demonstrations of the proposed path planning method are presented, which include various indoor and outdoor terrains and different vehicle models. The performance of the proposed method is also discussed in this section. Finally, in Section 6, the work is summarized and commented on.

\section{ML-SkiMap Generation}
\label{sec:Variable_resolution_point_cloud_in_ML-SkiMap}

For unstructured terrains, traditional representations use gridded Digital Elevation Maps (DEMs) \cite{chen2025learning} or curve-fitted surface patches \cite{zhang2025driving, zhang2024high}, which face a fundamental trade-off between terrain fidelity and computational efficiency. High-resolution terrain models impose significant burdens on data storage, processing, and transmission \cite{liu2011effects}, while coarse representations sacrifice geometric accuracy essential for safe navigation. This challenge is particularly acute for point-cloud-based planning methods \cite{krusi2017driving}, where dense point clouds drastically increase computation time due to the sheer volume of data requiring traversability assessment. To address this limitation, we employ a multi-resolution representation using ML-SkiMap, which maintains terrain detail where needed while reducing data density in geometrically simple regions, thereby achieving computational efficiency without compromising planning safety.

\subsection{ML-SkiMap}\label{sec:ml-skimap}
SkiMap is a mapping framework proposed in \cite{de2017skimap}, which is specifically designed to store sparse point clouds in 3D space. It has the combined advantages of a 3D grid in computational complexity and of an Octree in storage size. The data is organized using a SkipList tree with only the occupied cells stored. Besides, the SkipList tree provides efficient access (with calculation complexity of $O(log(n))$ ) to any voxel and the 3D points inside.

\begin{figure}[t]
\centering
     \begin{subfigure}[t]{0.45\textwidth}
         \centering
         \includegraphics[width=0.8\textwidth]{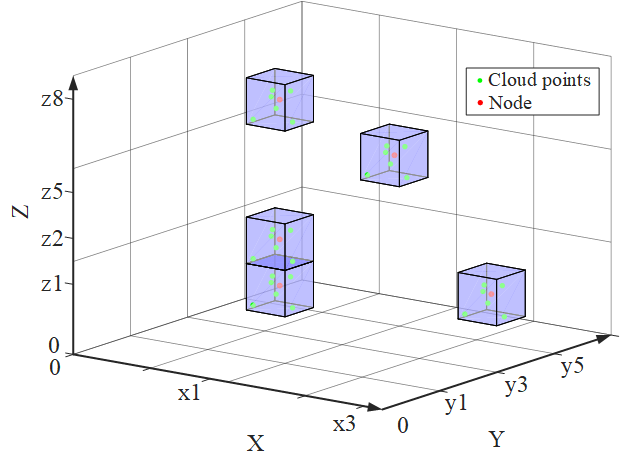}
         \caption{}
         \label{fig:fig1-a}
     \end{subfigure}
     \begin{subfigure}[t]{0.45\textwidth}
         \centering
         \includegraphics[width=\textwidth]{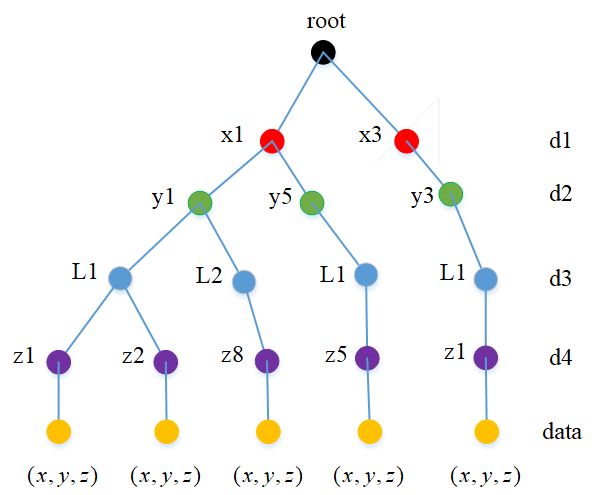}
         \caption{}
         \label{fig:fig1-b}
     \end{subfigure}
     \caption{Structure of the multi-layered Skimap. }     \label{fig1}
\end{figure}

Unlike the traditional path planning method, to treat the terrain as a single-layer surface, we deal with much more complex structures such as jungles, tunnels, and multi-layer buildings. Under such environments, it is essential to differentiate their vertical structures when evaluating the traversability of a UGV. The structure of the ML- SkiMap is shown in Figure \ref{fig1}. SkiMap divides the 3D space with quantized $(X, Y, Z)$ coordinates and provides sequentially arranged voxels, {as shown in Figure \ref{fig1}(a)}. For autonomous navigation, the square cell length $d$ for the map grid is suggested to be approximately the radius of the vehicle wheels so that the path planning does not give unreasonable decisions due to the lack of terrain information at places with low data density. 

The primary distinction of the ML-SkiMap from the original SkiMap structure is the addition of an extra layer that encodes the vertical levels of the cells. This layer is situated between the \( X \)- and \( Y \)-coordinate layers and the \( Z \)-coordinate layer. The cells in the \( Z \)-direction are partitioned into distinct groups that represent vertical structural information at a specific \( (X, Y) \) location. When the distance between adjacent voxels in the \( Z \)-direction exceeds a threshold \( h_L \), they are assigned to different levels. The value of \( h_L \), a user-defined parameter, is typically related to the height of the UGV, $H_R$. It is suggested to be set as 
\begin{equation}
h_L \geq \alpha \cdot H_R, \quad \alpha \in [1.5, 3.0]
\label{eq:hL_robot}
\end{equation}
where $\alpha$ is a safety factor. The lower bound ensures the robot can physically fit between levels, while the upper bound prevents over-segmentation of continuous sloped terrain. As highlighted by Krusi et al. \cite{krusi2017driving}, path planning using such a multi-level map can be regarded as an intermediate stage between 2.5D and full 3D path planning.

The ML-SkiMap tree organizes voxel information and corresponding points across five hierarchical layers, as shown in Figure \ref{fig1}(b). At depth \( d_1 \), nodes store the occupied \( X \)-indices, while depth \( d_2 \) contains the occupied \( Y \)-indices for the current index. Depth \( d_3 \) stores the level indices corresponding to the current \( (X, Y) \) cell, and at depth \( d_4 \), the tree records the \( Z \)-indices for the voxels at the current level. Finally, at depth \( d_5 \), nodes hold the points within the current \( (X, Y, Z) \) voxel. Since all indices from \( d_1 \) to \( d_4 \) are sorted, surface voxels and their corresponding points can be easily identified at any \( (X, Y) \) coordinate. These are represented by the first elements at depths \( d_4 \) and \( d_5 \), respectively.

\subsection{Variable resolution map}
The 3D space is divided into voxels $V_{i,j,k}^{l}$ stored in the ML-SkiMap, where the $i$, $j$, $k$ are the index of the voxel in $X$, $Y$, $Z$ coordinates and $l$ is the level index in the current $(X,Y)$ grid cell. The surface curvature calculation method is estimated numerically for the voxel $V_{i,j,k}^{l}$ based on a principal component analysis (PCA) process by calculating the eigenvectors ${}^m \mathbf{v}_{i,j,k}^{l}$ and eigenvalues  ${}^m \lambda_{i,j,k}^{l}$, as shown in Eq. (2), of a covariance matrix  $C_{i,j,k}^{l}$ defined in Eq.~\ref{eq1} \cite{tang2019autonomous}.

\begin{equation}\label{eq1}
    C_{i,j,k}^{l} = \frac{1}{n_{i,j,k}^{l}} \sum_{r=1}^{n_{i,j,k}^{l}} (\mathbf{q}_r - \mathbf{O}_{i,j,k}^{l}) (\mathbf{q}_r - \mathbf{O}_{i,j,k}^{l})^T,
\end{equation}

where $n_{i,j,k}^{l}$ is the number of the points in $V_{i,j,k}^{l}$, $\mathbf{q}_r$ is the $r^{th}$ 3D point in $\mathbf{V}_{i,j,k}^{l}$, $\mathbf{O}_{i,j,k}^{l}$ is the center of $\mathbf{V}_{i,j,k}^{l}$.

\begin{equation}\label{eq2}
    C_{i,j,k}^{l} \cdot {}^m \mathbf{v}_{i,j,k}^{l} = {}^m \lambda_{i,j,k}^{l} \cdot {}^m \mathbf{v}_{i,j,k}^{l}, \quad m \in {0,1,2}
\end{equation}

The surface curvature at $V_{i,j,k}^{l}$ is then estimated as 
\begin{equation}\label{eq3}
    {Curv}_{i,j,k}^{l} = \frac{{}^0 \lambda_{i,j,k}^{l}}{{}^0 \lambda_{i,j,k}^{l} + {}^1 \lambda_{i,j,k}^{l} + {}^2 \lambda_{i,j,k}^{l}},
\end{equation}

where ${}^0 \lambda_{i,j,k}^{l}$ is the largest eigenvalue of $C_{i,j,k}^{l}$.  

\begin{figure}[t]
\centering
     \begin{subfigure}[t]{0.45\textwidth}
         \centering
         \includegraphics[width=0.8\textwidth]{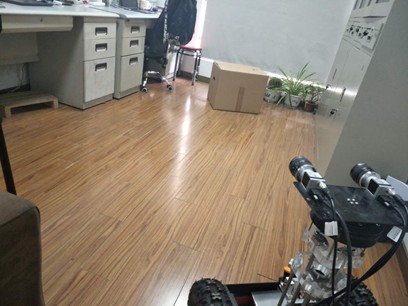}
         \caption{}
         \label{fig:fig2-a}
     \end{subfigure}
     \begin{subfigure}[t]{0.45\textwidth}
         \centering
         \includegraphics[width=\textwidth]{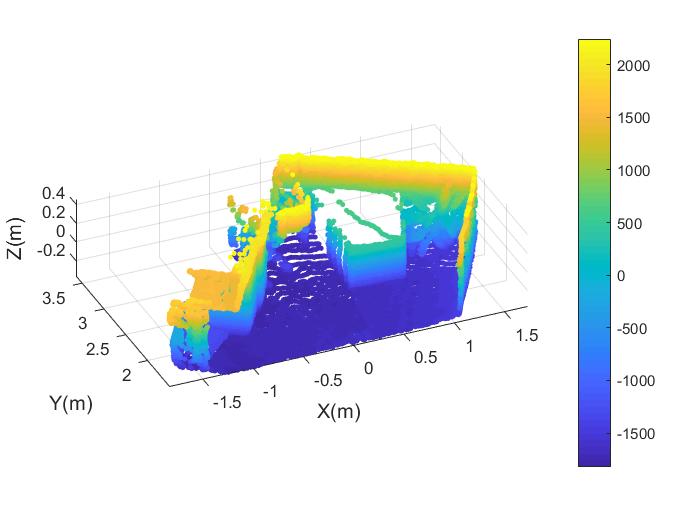}
         \caption{}
         \label{fig:fig2-b2}
     \end{subfigure}
     \\
     \begin{subfigure}[t]{0.45\textwidth}
         \centering
         \includegraphics[width=\textwidth]{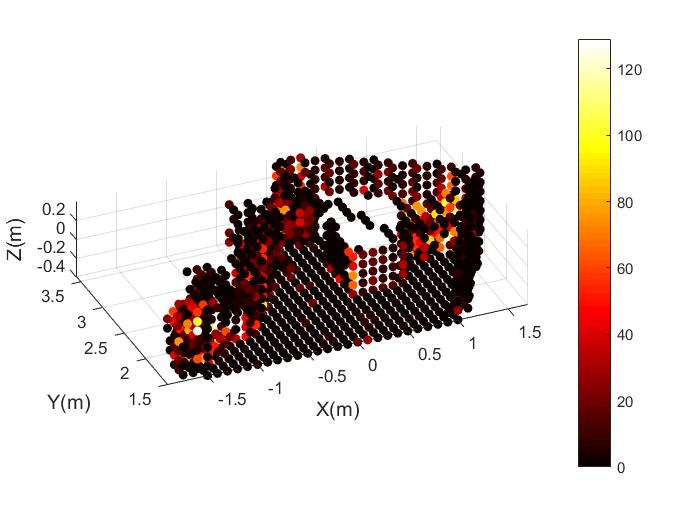}
         \caption{}
         \label{fig:fig2-c}
     \end{subfigure}
     \begin{subfigure}[t]{0.45\textwidth}
         \centering
         \includegraphics[width=\textwidth]{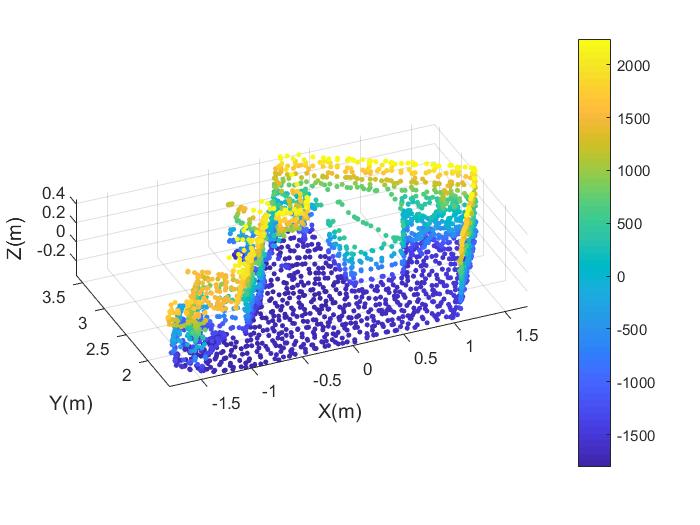}
         \caption{}
         \label{fig:fig2-d}
     \end{subfigure}
     \caption{Point cloud simplification process. (a) Test scene. (b) Original dense point cloud acquired from the highlighted area. (c) Curvature distribution map computed from local surface geometry. (d) Variable-resolution map after simplification. }     \label{fig2}
\end{figure}

The number of the data points $N_{i,j,k}^{l}$ to be preserved in  $V_{i,j,k}^{l}$ is then determined by the local curvature ${Curv}_{i,j,k}^{l}$, as in Eq.~\ref{eq4}. $N_{i,j,k}^{l}$ points in $V_{i,j,k}^{l}$ are randomly chosen from the original point set and stored in the corresponding tree node in the $d_5$ layer of the ML-SkiMap. 

\begin{equation}\label{eq4}
    N_{i,j,k}^{l} = a \times ({Curv}_{i,j,k}^{l})^b + c,
\end{equation}
where $a>0, b>0, c>0$ are parameters to be tuned by the users. The parameter $c$ specifies the minimum number of points retained in a voxel when the curvature is very small or equal to zero, effectively establishing a baseline point density. The parameters $a$ and $b$ control the shape of the curvature-to-point-count mapping function, as illustrated in Figure~\ref{fig2}(a) and (b). In practice, $c$ can be chosen as a small integer to minimize overall data size while maintaining sufficient coverage in low-curvature regions. The exponent $b$ is typically set larger than 1 to enhance the contrast ratio between high-curvature and low-curvature regions, allocating more points to geometrically complex areas. The scaling factor $a$ controls the maximum allowable point count within a voxel and should be increased when the environment contains substantial vertical features or complex three-dimensional structures that require denser representation. A more detailed explanation of parameter tuning can be found in our previous work \cite{tang2019autonomous}.

Figure \ref{fig2} shows an example for the point cloud simplification. The area under the light blue mask in Figure 2(a) is scanned, and the original dense point cloud is shown in Figure 2(b). The number of the point is 28579 for such a small area.

Choose the cell size as $100$ $\textnormal{mm}^3$ in the ML-SkiMap and implement the curvature calculation method on each voxel. The resulted curvature distribution is shown in Figure \ref{fig2}(c). The curvature represents the local complexity of the object shapes. For flat areas, such as the walls and the ground, the curvature value is small. For the parts with distinguished features, such as the edge of the box, the subtle shape of the desks and the chair, and the complex plant shape, the curvatures tend to be large. Applying Eq. \ref{eq4} for variant resolution map generation with coefficients $a_1=900, b_1=3, c_1=1$, the result in Figure \ref{fig2}(c) shows that the resolution for the flat area is largely decreased, and the complex shapes are preserved. The number of the point is decreased to be 2771, which is only 9.6\% of the original map. The variable resolution ML-SkiMap slims the data set and stores them in a more structured way. 

\section{Weighted traversability graph}\label{sec:Weighted_traversability_graph}

\subsection{Relative work discussion}
Traversability analysis is the prerequisite for path planning on 3D terrains. Traditionally, the point clouds are further used to compute a height or voxel map, such as simplified height patches \cite{gu2008rapid}, ITM \cite{papadakis20123d}, or DEM with some fitting techniques \cite{papadakis20123d}, to provide processed environment information for further traversability analysis. However, it was pointed out that the point cloud data, compared to other artificial maps, is easier to obtain and maintain by avoiding any supplementary terrain reconstruction process \cite{krusi2017driving}. The majority of the traversability analysis methodologies used only geometric information of the terrain \cite{papadakis2013terrain, leininger2024gaussian}. A traversability index at a specific location quantified the traversing safety based on the terrain's slope and roughness \cite{gu2010diverse}. Often the local 3D points were fitted as a plane, the norm of which functions as slope estimation, and the residue of the fitness represented the roughness. Other considerations, such as the height difference, the pitch angle, and the roll angle, were also added into the traversability cost formula in some studies \cite{bellone2013unevenness,kubota2001path}. 

However, the driving safety of any ground robot is also affected by the vehicle's properties besides the terrain conditions \cite{krusi2017driving}. If the vehicle's kinematic constraints and interactive relationship with the terrain were dropped out as in the methods mentioned above, the resulted path sometimes could fail \cite{reddy2016computing,reina20143d}. For example, without considering the robot's width, the planned path might make the robot travel through a too narrow channel \cite{rekleitis2007over}. A few works solved the problem by introducing the vehicle's properties in traversability assessments with static or dynamic analysis \cite{vandapel2006unmanned,ishigami2011path}. They functioned well in addressing the vehicle-terrain interaction relationship, while the models' fidelity must be carefully controlled to keep the algorithm from prohibitively time-consuming. In our previous work, we proposed a traversability assessment method with 2D grid taking into consideration of the geometric relationship between the point cloud and the vehicle \cite{tang2019autonomous}. In such a way, the driving safety is not only determined by the geometry of the ground surface but also their interactive relationship when the vehicle is placed on the terrain at a specific location in a certain direction. However, it assumed a single-layer ground surface, which limits its applications for more complex environments with vertically overlapping features. 

In this section, we improve the single-layer analysis to be an ML-SkiMap version, which combines the vehicle-environment interacted analysis method in \cite{tang2019autonomous} with the map structure presented in Section \ref{sec:Variable_resolution_point_cloud_in_ML-SkiMap}. The driving safety of a grid on the x-y plane is evaluated at multiple layers, and a connectivity network is used to represent the terrain structure. Together, they provide a weighted traversability graph (WTG). 

\subsection{WTG definition}
WTG is a directed graph that encodes traversability costs of multi-level terrain. Let the environment be discretized into a 2D grid with cells $\mathcal{A} = \{a_{i,j} 
\mid i \in [1, N_x], j \in [1, N_y]\}$, where each cell $a_{i,j}$ has size $d \times d$ 
in the horizontal plane. For each cell $a_{i,j}$, the ML-SkiMap identifies $L_{i,j}$ 
distinct vertical levels, indexed by $l \in [1, L_{i,j}]$, where surfaces are 
separated by a minimum vertical distance $h_L$ (Section~\ref{sec:ml-skimap}). Each level $l$ at position $(i,j)$ is represented by its top surface voxel:
\begin{equation}
V_{i,j,k}^l \in \mathbb{R}^3, \quad k = \arg\max_{k'} \{z \mid (i,j,k') \in \text{Level } l\}
\end{equation}
where $k$ denotes the discrete $z$-index of the topmost voxel in level $l$, as ground 
vehicles interact only with the upper surface of each terrain layer.

\textbf{Node Set:} The WTG is defined as $G = (\mathcal{V}, \mathcal{E}, W)$, where 
the node set is:
\begin{equation}
\mathcal{V} = \{v_{i,j}^l \mid a_{i,j} \in \mathcal{A}, l \in [1, L_{i,j}]\}.
\end{equation}
Each node $v_{i,j}^l$ corresponds to the $l$-th level surface at horizontal cell 
$(i,j)$, with 3D position given by the centroid of $V_{i,j,k}^l$.

\textbf{Edge Set:} The directed edge set $\mathcal{E} \subseteq \mathcal{V} \times 
\mathcal{V}$ consists of two types:

\noindent\textit{(1) Intra-level edges} (horizontal traversal within the same level):
\begin{equation}
\mathcal{E}_{\text{intra}} = \{(v_{i,j}^l, v_{m,n}^{l'}) \mid (m,n) \in \mathcal{N}(i,j), 
\text{ and } C_{i,j \to m,n}^{l \to l'} = 1\},
\end{equation}
where $\mathcal{N}(i,j) = \{(i\pm1, j\pm1), (i\pm1, j), (i, j\pm1)\}$ denotes the 
8-connected neighborhood, and $C_{i,j \to m,n}^{l \to l'}$ is the connectivity indicator 
(defined below).

\noindent\textit{(2) Inter-level edges} (vertical transitions, e.g., ramps, stairs):
\begin{equation}
\mathcal{E}_{\text{inter}} = \{(v_{i,j}^l, v_{i,j}^{l'}) \mid l \neq l', 
\text{ and } |z_l - z_{l'}| \leq h_{\text{climb}}\},
\end{equation}
where $h_{\text{climb}}$ is the maximum vertical step height the robot can traverse 
(e.g., stair height). Inter-level edges at the same $(i,j)$ cell enable vertical 
transitions when terrain features (ramps, steps) connect multiple levels locally.

\textbf{Connectivity Criterion:} The binary connectivity indicator $C_{i,j \to m,n}^{l 
\to l'} \in \{0, 1\}$ determines whether nodes $v_{i,j}^l$ and $v_{m,n}^{l'}$ are 
connected:
\begin{equation}
C_{i,j \to m,n}^{l \to l'} = 
\begin{cases}
1, & \text{if } |k_l - k_{l'}| \leq c_{\max} \\
0, & \text{otherwise}
\end{cases}
\label{eq:connectivity}
\end{equation}
where $k_l$ and $k_{l'}$ are the $z$-indices of the top surface voxels at levels $l$ 
and $l'$, respectively, and $c_{\max} = \lceil h_{\text{climb}} / d_z \rceil$ is the 
maximum allowable vertical step in voxel units, with $d_z$ being the voxel height. This 
criterion ensures that only geometrically feasible transitions (within the robot's 
climbing capability) are permitted.

\begin{figure} [H]
    \centering
    \includegraphics[height=2.50in]{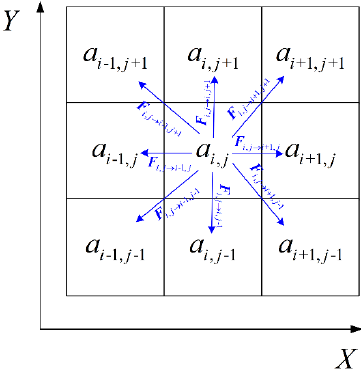}
    \hspace{.1in}
    \caption{Definition of the traversability vector $F_{i,j}$.}
    \label{fig3}
\end{figure}

\textbf{Edge Weights:} For each directed edge $e = (v_{i,j}^l, v_{m,n}^{l'}) \in 
\mathcal{E}$, the traversability weight is:
\begin{equation}
W(e) = F_{i,j \to m,n}^l \in \mathbb{R}_{\geq 0},
\label{eq:edge_weight}
\end{equation}
where $F_{i,j \to m,n}^l$ is the directional traversability cost from cell $(i,j)$ at 
level $l$ toward neighbor $(m,n)$, computed from terrain analysis (Section~\ref{sec:traversability_assessment}). The 
traversability vector at node $v_{i,j}^l$ is:
\begin{equation}
\mathbf{F}_{i,j}^l = [F_{i,j \to m,n}^l]_{(m,n) \in \mathcal{N}(i,j)} \in \mathbb{R}^8,
\end{equation}
capturing the cost to traverse in each of the eight cardinal/diagonal directions 
$\phi \in \{0°, 45°, 90°, 135°, 180°, 225°, 270°, 315°\}$ (Figure~\ref{fig3}).

\begin{figure}[H]
\centering
     \begin{subfigure}[t]{0.165\textwidth}
         \centering
         \includegraphics[height=1.65in]{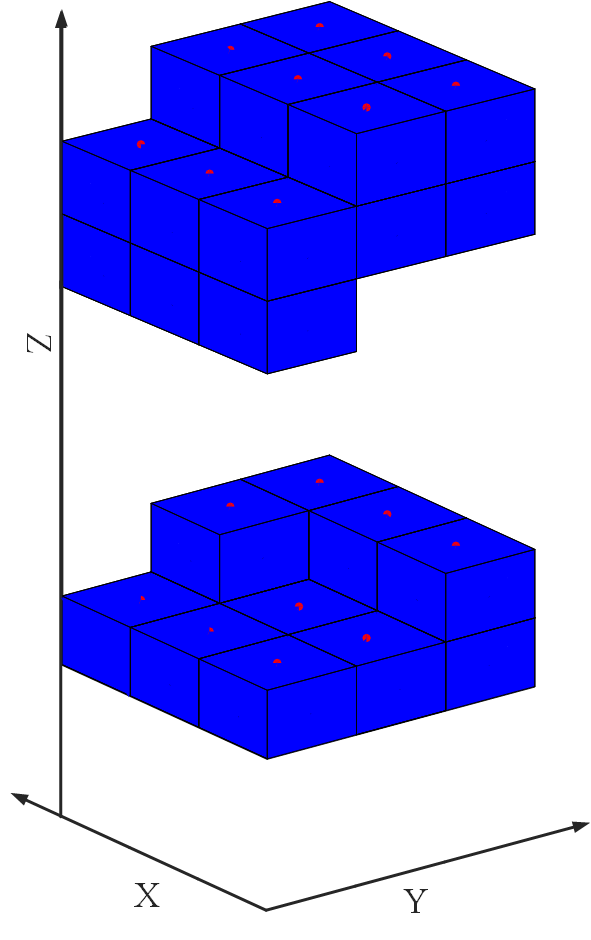}
         \caption{\bluetext{}}
         \label{fig:fig4-a}
     \end{subfigure}
     \begin{subfigure}[t]{0.165\textwidth}
         \centering
         \includegraphics[height=1.65in]{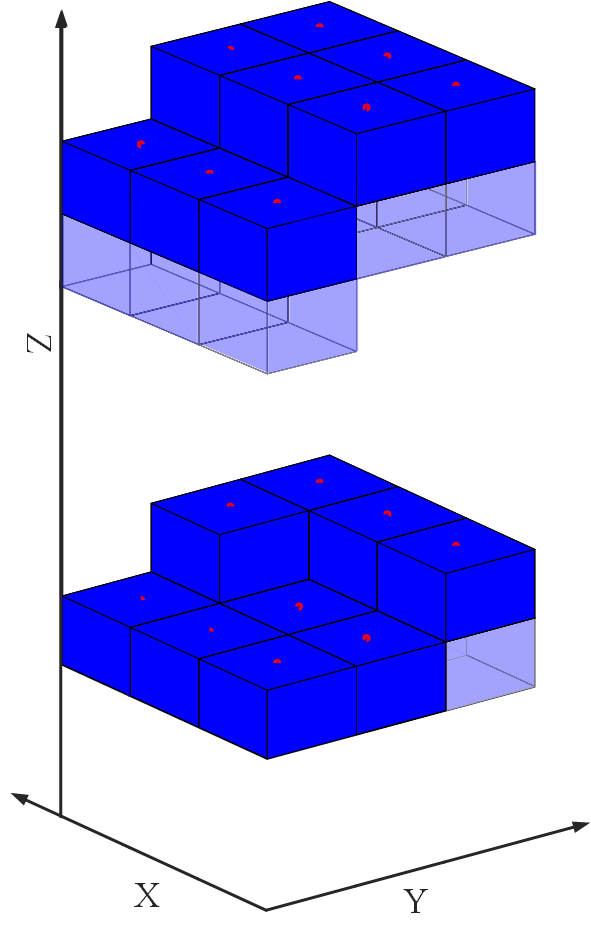}
         \caption{\bluetext{}}
         \label{fig:fig4-b}
     \end{subfigure}
     \begin{subfigure}[t]{0.165\textwidth}
         \centering
         \includegraphics[height=1.9in]{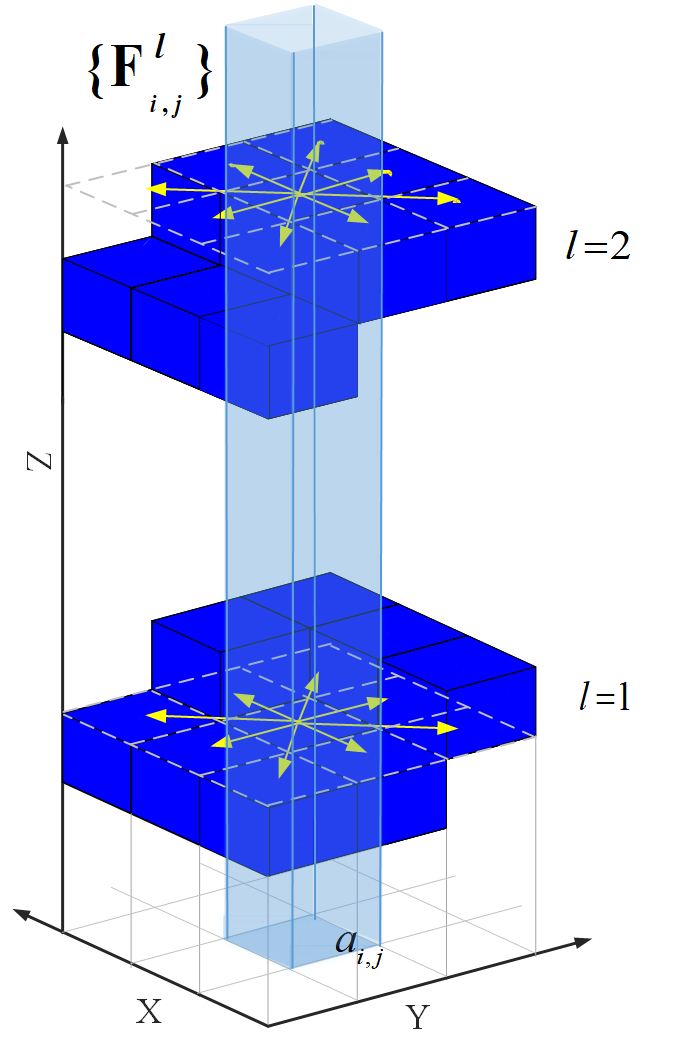}
         \caption{\bluetext{}}
         \label{fig:fig4-c}
     \end{subfigure}
     \begin{subfigure}[t]{0.165\textwidth}
         \centering
         \includegraphics[height=1.9in]{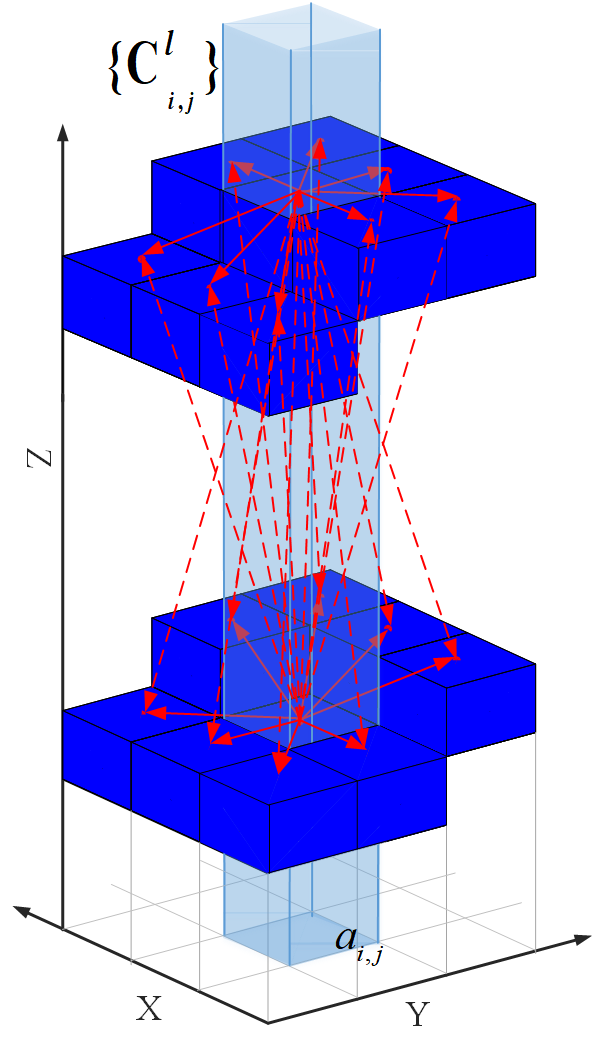}
         \caption{\bluetext{}}
         \label{fig:fig4-d}
     \end{subfigure}
     \begin{subfigure}[t]{0.165\textwidth}
         \centering
         \includegraphics[height=1.9in]{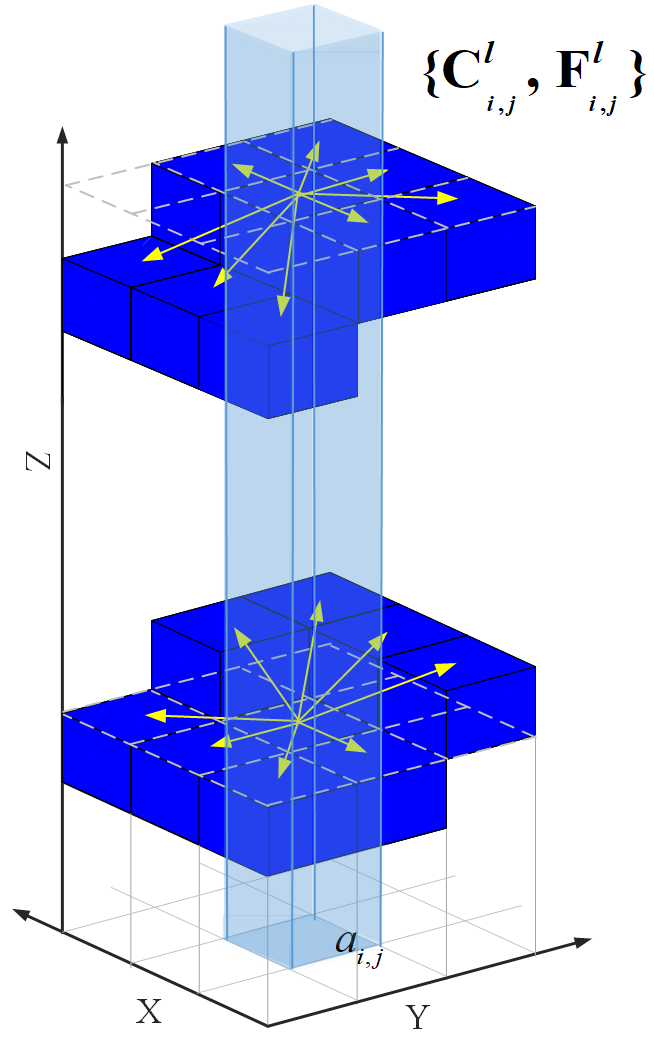}
         \caption{\bluetext{}}
         \label{fig:fig4-e}
     \end{subfigure}
    \caption{Weighted traversability graph construction. (a) the local ML-SkiMap voxels. (b) the extraction of the local surface voxels around some $a_{i.j}$. (c) connectivity analysis of the multi-level surfaces around $a_{i.j}$. (d) traversability analysis at $a_{i.j}$ in 8 directions on two levels. (e) combining results of $C_{i,j}^l$ and $F_{i,j}^l$ to get the WTG.}     \label{fig4}
\end{figure}

The complete WTG is constructed by:
\begin{enumerate}
    \item \textbf{Surface extraction}: For each $(i,j,l)$, identify top surface voxel 
    $V_{i,j,k}^l$ (Figure~\ref{fig4}(b)).
    \item \textbf{Traversability computation}: Calculate $\mathbf{F}_{i,j}^l$ using 
    vehicle-terrain interaction analysis (Section~\ref{sec:traversability_assessment}).
    \item \textbf{Connectivity evaluation}: For all neighbor pairs and level 
    combinations, compute $C_{i,j \to m,n}^{l \to l'}$ via Eq.~\ref{eq:connectivity} 
    (Figure~\ref{fig4}(c)).
    \item \textbf{Graph assembly}: Add edge $(v_{i,j}^l, v_{m,n}^{l'})$ with weight 
    $W = F_{i,j \to m,n}^l$ if $C_{i,j \to m,n}^{l \to l'} = 1$ (Figure~\ref{fig4}(e)).
\end{enumerate}
The resulting graph $G$ encodes all traversable paths in the multi-level environment, 
enabling standard graph search algorithms (A*, Dijkstra) to find globally optimal paths 
that may span multiple levels.

\subsection{Calculation of the traversability weights}\label{sec:traversability_assessment}
Traversability assessment is to consider different hazard situations for a vehicle at a specific location and pose. The method in this article to calculate $F_{i,j}^l$ is similar to \cite{tang2019autonomous}, with some modifications to accommodate the multi-level map structures. 

\subsubsection{Tip-over and slip hazards}
Being tipped over or slipping at steep slopes are two main hazards that need to be addressed when driving through unknown territories. Most of the previous researches, as stated above, use only the estimation of the terrain slope and roughness to evaluate the possibility for the two hazards to happen, which is not sufficient when the terrain is much complicated. In the limited number of literature incorporating vehicle properties, generally, only static cases are considered \cite{bellone2013unevenness,albrecht2014curvature}. An accurate dynamic model is hard to obtain because it not only involves the vehicle dynamics, which is already complex for certain systems, but also highly related to the complicated wheel-terrain interaction mechanism. Therefore, we adopt static analysis in this article. 

\begin{figure}[h]
    \centering
    \includegraphics[height=1.8in]{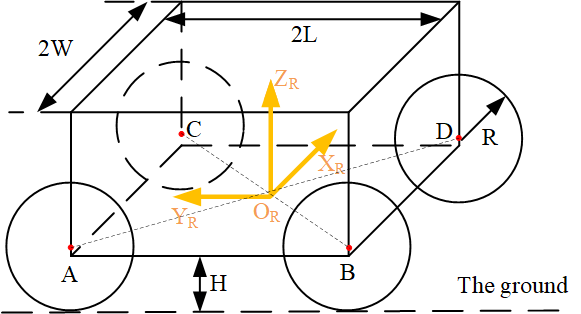}
    \caption{A simplified geometric model of the vehicle.}
    \label{fig5}
\end{figure}

The vehicle is simplified as shown in Figure \ref{fig5}, with the critical geometric parameters as the distance between the left and right wheels $2W$, the distance between the front and rear wheels $2L$, the radius of the wheel  $R$, and the height of the chassis $H$. The vehicle coordinate system is defined to have the origin at the geometric center of the four wheels. Then the centers of the four wheels on the x-y plane is $C_A=(-W,L,0)$, $C_B=(-W,-L,0)$, $C_C=(W,L,0)$,  $C_D=(W,-L,0)$ in the vehicle coordinate system.

To calculate the projection of the vehicle on the x-y plane, the centers of the four wheels on the terrain map in the x-y plane is obtained using the formula in Eq. \ref{eq6}, where $C$ is the coordinate value of a specific point in the vehicle coordinate system, $M$ is the corresponding coordinate value in the terrain map coordinate system, $\phi$ is the angel of the placement direction and $T=(T_x,T_y,T_z)$ is the translation vector from the current vehicle coordinate system to the terrain map coordinate system. 

\begin{equation}\label{eq6}
    \mathbf{M} =  
    \begin{bmatrix}
    cos \phi & -sin \phi & 0 \\
,    sin \phi & cos \phi & 0 \\
    0 & 0 & 1
    \end{bmatrix} \mathbf{C} +\mathbf{T}.
\end{equation}

\begin{figure}[h]
    \centering
    \includegraphics[height=2.5in]{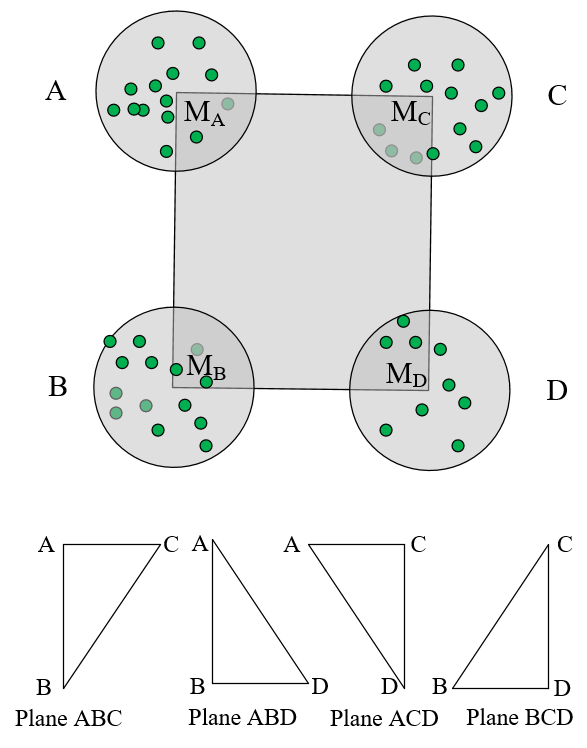}
    \hspace{.2in}
    \caption{Footprints of the vehicle and the four corresponding fitted planes to estimate the vehicle pose.}
    \label{fig6}
\end{figure}

Correspondingly, the footprints of the four wheels on the terrain map are defined as the four circles with a radius $R$ of centered at $\mathbf{M}_{\mathbf{A}}$, $\mathbf{M}_{\mathbf{B}}$, $\mathbf{M}_{\mathbf{C}}$, $\mathbf{M}_{\mathbf{D}}$, respectively (Figure \ref{fig6}). When the vehicle is driving on uneven terrains, technically, the four wheels can not be guaranteed to be all on the ground simultaneously. Without rigorous dynamical analysis, it is impossible to determine which three wheels would be on the ground at the current location. Therefore, we choose to consider all four possibilities. That is, as shown in Figure \ref{fig6}, to examine all the four cases when wheels ABC, ABD, ACD, BCD are on the ground, respectively. 

\begin{figure}[h]
    \centering
    \includegraphics[height=2.3in]{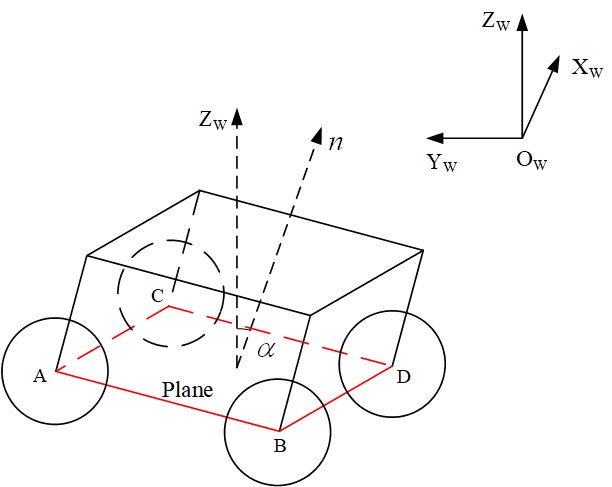}
    \hspace{.2in}
    \caption{The tilt angle of the vehicle.}
    \label{fig7}
\end{figure}

For each case, the terrain point clouds of the corresponding footprints are fit as a plane. These 3D points are indexed by the ML-SkiMap vector $x,y,l.z$, where $x,y$ is the coordinates of the wheel footprint projected on the $x,y$ plane, $l$ is the layer closest to the vehicle coordinate system and $z$ is the height value of the surface voxel in layer $l$. The corresponding planes are obtained by the PCA technique, and the normal vector $\mathbf{n}_w, w=1,2,3,4$ of the fitted plane is the eigenvector corresponding to the minimal eigenvalue of the resulting covariance matrix $\mathbf{C}_w$, as defined in Eq.~\ref{eq7}. The angle $\alpha_w$ between the vertical vector and $\mathbf{n}_w$ is the estimation of the inclination of the vehicle chassis, as illustrated in Figure~\ref{fig7} and formulated in Eq.~\ref{eq9}. The corresponding plane equation is obtained using Eq. \ref{eq9}. As the force-angle relationship is stated in previous literature \cite{rekleitis2007over,nash2007theta}, the larger the angle $\alpha_w$, the more dangerous the vehicle is at the current location and driving direction. In this article, we simplify the constraints to be the maximum chassis tilt angle. For vehicles with significantly different allowable roll and pitch angles, the constraints can be adapted by using the head angle $\phi$ to separately calculate the roll and pitch angles.

\begin{equation}\label{eq7}
    \mathbf{C}_w = \frac{1}{k_w} \sum_{i=1}^{k_w} (\mathbf{q}_{iw} - \Bar{\mathbf{q}}_w)(\mathbf{q}_{iw} - \Bar{\mathbf{q}}_w^T),
\end{equation}
where $\mathbf{q}_{iw}$ belongs to the point cloud set of one of the footprint cases {ABC}, {ABD}, {ACD} and {BCD}, $k_{w}$ is the number of the points in the set, and $\Bar{\mathbf{q}}_{w}$ is the average value of the point set. 

\begin{equation}\label{eq8}
    \alpha_w = \mathrm{arcos} (\frac{\mathbf{n}_w \cdot \mathbf{Z}_W}{||n_w||_2}),
\end{equation}

\begin{equation}\label{eq9}
    \mathbf{n}_w \cdot 
    \begin{bmatrix}
    x\\y\\z
    \end{bmatrix}
    + \mathbf{n}_w \cdot \bar{\mathbf{q}}_w = \mathbf{0},
\end{equation}

\subsubsection{Treatment of the occlusion}\label{sec_occlusion}
\begin{figure}[h]
    \centering
    \includegraphics[height=2.7in]{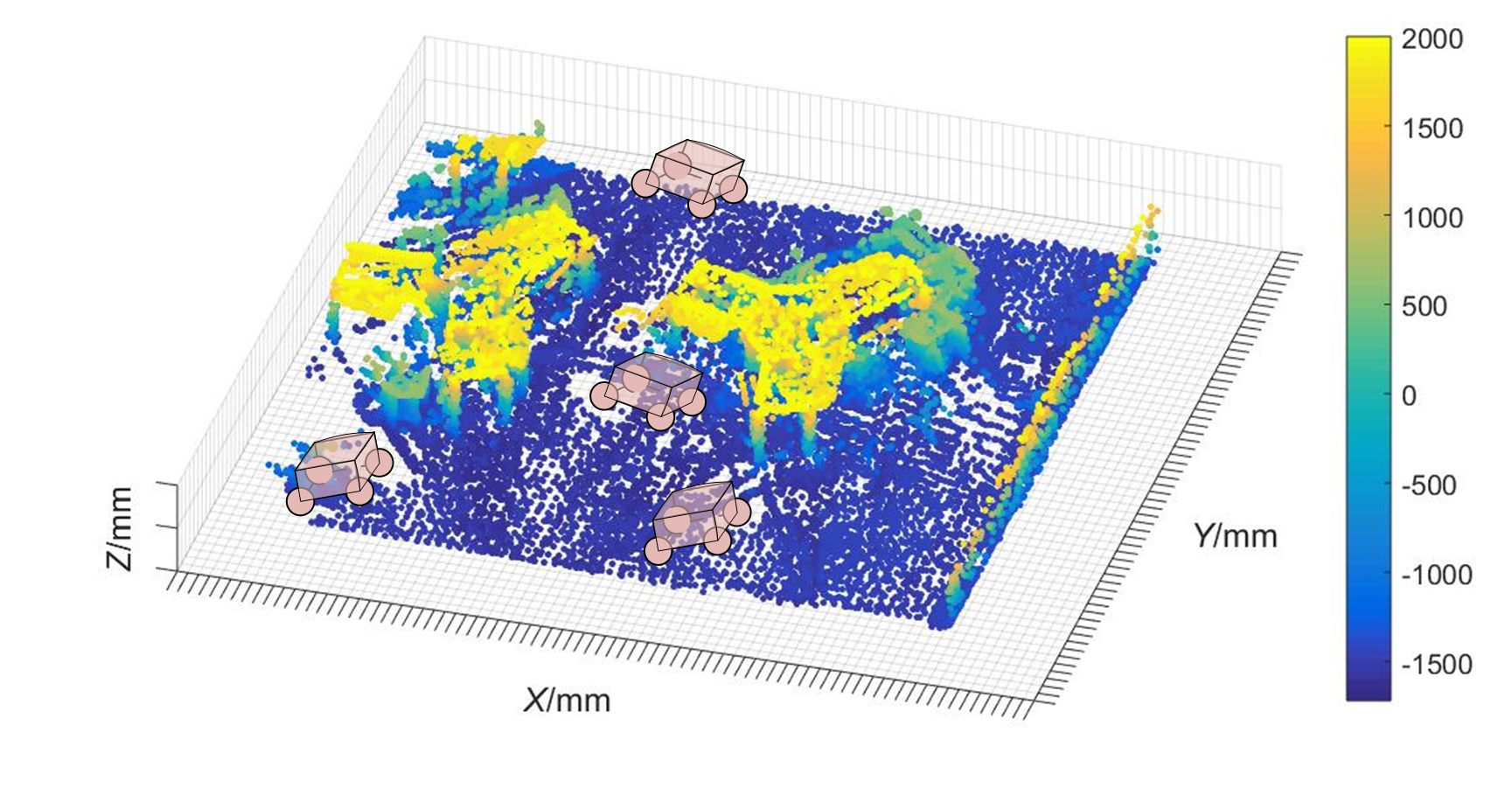}
    \hspace{.2in}
    \caption{Cases when the vehicle is at the occlusion areas.}
    \label{fig8}
\end{figure}

Limited by the capability of the onboard sensors, there are always areas where the terrain information is missing due to occlusions, as shown in Figure \ref{fig8}. An x-y cell is defined as an occlusion area when there is no point cloud within it. Due to the insufficiency of the terrain information in these areas, the traversability of the vehicle at the location is hard to determine. We use a conservative strategy in this article to treat the problem. When any of the footprints of the four wheels are at the area of the occlusion, as shown in Figure \ref{fig8}, the corresponding pose of the vehicle is considered to be dangerous. This can successfully avoid any wheel falls in small holes or narrow abysses, which are usually present as blank areas in point clouds. 

\subsubsection{Chassis collision hazards}
\begin{figure}[h]
    \centering
    \includegraphics[height=1.50in]{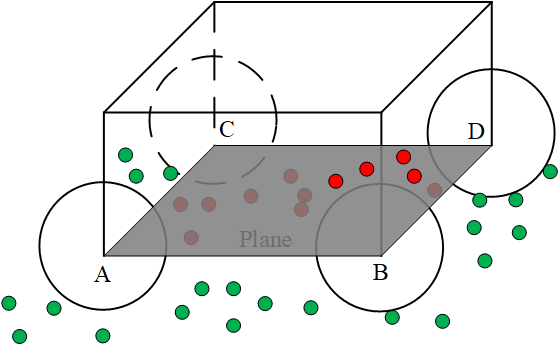}
    \hspace{.2in}
    \caption{Chassis collides with obstacle points.}
    \label{fig9}
\end{figure}

Another incident that could cause damages to the vehicle is when the chassis collides with the obstacles on the ground. This could happen even when the four wheels sit safely on relatively flat ground without danger of tip-over or slippery, as shown in Figure \ref{fig9}.

The danger of chassis collision can be checked by calculating the relative relationship between the chassis plane and the terrain surface points. The terrain surface refers to the surface of the layer below, which is closest to the origin of the vehicle coordinate system. The chassis is assumed to be parallel with the footprint planes. For each of the footprint cases {ABC}, {ABD}, {ACD}, {BCD}, the chassis plane is obtained by making a plane shift $H$ to Eq. \ref{eq9}, which is given by:

\begin{equation}\label{eq10}
    \mathbf{n}_w \cdot 
    \begin{bmatrix}
    x\\y\\z
    \end{bmatrix}
    - \mathbf{n}_w \cdot \bar{\mathbf{q}}_w + H = \mathbf{0}.
\end{equation}

Denote the terrain point clouds within the square $M_A M_B M_C M_D$ as $P{MS}$, and for each point $p \in P_{MS}$, it collides with the chassis when  
\begin{equation}\label{eq11}
    \mathbf{n}_w \cdot \mathbf{p}  >   -\mathbf{n}_w \cdot \bar{\mathbf{q}}_w + H.
\end{equation}

\subsubsection{Traversability cost calculation}
Combined with all the considerations listed above, $\mathbf{F}_{i,j}^{l}$ can be obtained. When the hazards of falling in an unknown area, flip over, or chassis collision could happen, the traversability cost of the corresponding pose is set as a very large value $\eta$ to prevent vehicles from passing by. Otherwise, it indicates the current pose on the point cloud is feasible, and the traveling cost is set as the tangent of the tilt angle. The corresponding algorithm is shown below:

\begin{algorithm} [H]
\caption{$\mathbf{F}_{i,j}^{l}$ calculation}\label{alg1}
\textbf{Input:} ML-SkiMap \\
\textbf{Output:} $\mathbf{F}_{i,j}^{l}$ on each level of all occupied x-y cell
\begin{algorithmic} 
\For {each $x$ in index $i$}
    \For {each $y$ in index $j$ in current $i$}
        \For {each level $l$ }
            \For {each $\phi$ value }
                \If{any footprint of the four wheels has no points }
                    {$F_{i,j}(\phi)=\eta$}
                \ElsIf{ ($\mathbf{n}_w \cdot \mathbf{p}  > -\mathbf{n}_w \cdot \bar{\mathbf{q}}_w +H$)}
                    {$F_{i,j}(\phi)=\eta$}
                \Else
                    {\quad $F_{i,j}(\phi)=\mathrm{tan} \alpha_w$}
                \EndIf
            \EndFor  
        \EndFor
    \EndFor
\EndFor
\end{algorithmic}
\end{algorithm}

\subsubsection{Case Study for the WTG}
\begin{figure} [H]
    \centering
    \includegraphics[height=2.50in]{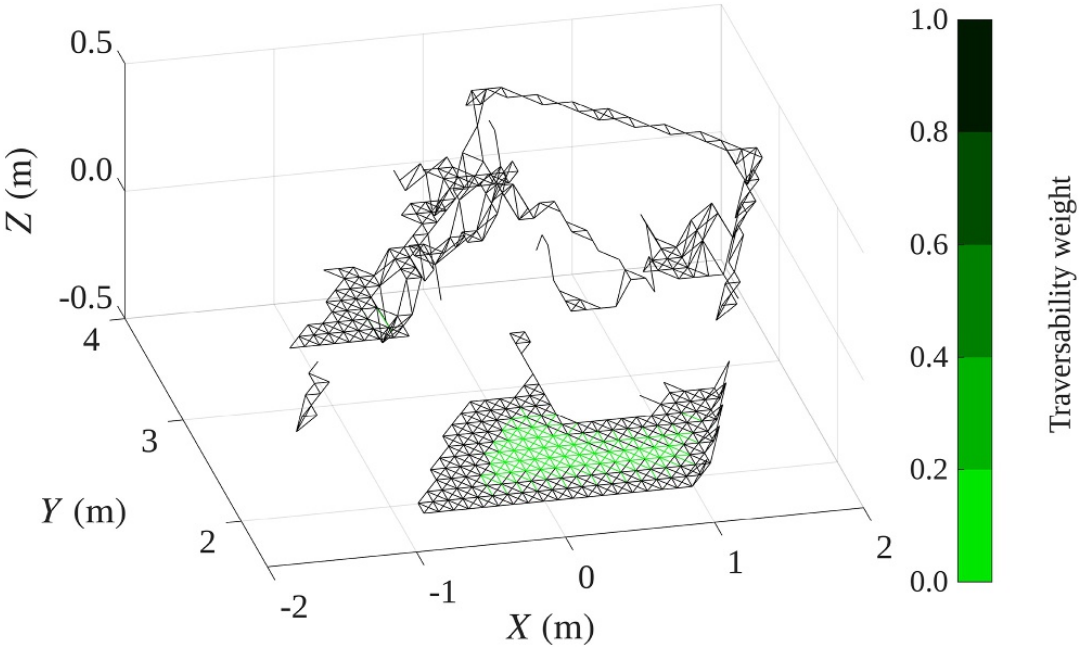}
    \hspace{.2in}
    \caption{The visualization of the WTG for the testing ground in Figure \protect\ref{fig3}}
    \label{fig10}
\end{figure}

The terrain map in Figure~\ref{fig2} is used to test the generation of the WTG. The surface voxels for each level are selected, and their connectivity with the surface voxels on the neighbor x-y cells is calculated. For the particular vehicle in the test, R=12cm, L=17.5cm, W=23.5 cm, H=9 cm. We set the maximum height difference of the two voxels $c_{max}$ as 10 cm. The resulted WTG is shown in Figure \ref{fig10}. The surface cells are extracted, and if there is a valid connection between two neighbor cells, they are connected by a line. The color of the line indicates the normalized traversability between the two. If it is not traversable, the color tends to be black; otherwise, it tends to be green. It should be pointed out that we only use one value of $\mathbf{F}_{i,j}^{l}$ on the same edge when drawing the WTG to the simplicity of visualization. Particularly in this graph, the in the driving directions of 0°, 45°, 90°, and 135° are used. The WTG in directions of 180°, 225°, 270° and 315° is similar and neglected in the following discussion.

It is obvious that the voxels of the same surface, such as the voxels belonging to the ground surfaces, are determined as connected. On the other hand, when there is a large height difference between two neighbor cells, such as at the edge of the desks and the ground, they are diagnosed as disconnected. The center area of the ground is green, meaning the vehicle can move in any direction without hazards. On the other hand, the WTG determines the boundary area of the ground as not traversable because the vehicle either impact the surrounding obstacles or step on unknown terrain. There is a small green area on the desk because the desk is just large enough to fit the vehicle in. However, the vehicle can never reach the ground from the desk because there are no connected cells between the two.

\section{Graph-based path search}\label{sec:Graph-based_path_search}
With the WTG obtained, the traversability network with weighted connections is established. In this case, graph search methods can be utilized for path planning tasks. The widely used methods for autonomous vehicles include A* search \cite{potiris2014terrain}, D* Search \cite{carsten2007global}, and Basic theta* \cite{nash2007theta}. The readers are referred to the review article \cite{janis2016path} for the commonly seen planning algorithms in complex environments.

We adopt the A* algorithm in this article, a typical heuristic search method with both considerations on calculation efficiency and path cost. The commonly seen A* algorithms are implemented on 2D or 3D occupancy maps. It is modified to deal with the WTG structure proposed in this article. The major difference comes from the calculation of the travel cost. Given a starting voxel and a target voxel, A* algorithm heuristically searches the neighbor voxels in WTG to extend its path at each step. In the 3D space, only the connected voxels would be considered as candidates. The corresponding travel cost $f$ in Eq. \ref{eq13} is a combination of the accumulated path cost $g_{m,n}^p$, the heuristic target distance $h_{m,n}^p$ and the current travel cost $\mathbf{F}_{i,j \rightarrow m,n}^{k}$. The travel cost at a candidate voxel $V_{m,n}^p$ is not only determined by itself, but also by its parent voxel $V_{i,j}^k$. 

\begin{equation}\label{eq13}
    f(V_{m,n}^p) = g_{m,n}^p + h_{m,n}^p + \lambda \cdot \mathbf{F}_{i,j \rightarrow m,n}^{k},
\end{equation}

where $V_{m,n}^p$ is an adjacent node to $V_{i,j}^k$; $f(V_{m,n}^p)$ is the estimated total travel cost of $V_{m,n}^p$; $g_{m,n}^p$ is the actual travel cost from the start voxel to $V_{m,n}^p$; $h_{m,n}^p$ is the heuristic estimated distance cost from $V_{m,n}^p$ to the goal (the Euclidean distance is used here); $\mathbf{F}_{i,j \rightarrow m,n}^{k}$ is the F-map value at $V_{i,j}^k$ in the direction from $k^{th}$ level cell of (i, j) to $p^{th}$ level cell of (m, n); $\lambda$ is a constant value to balance the distance cost and the traversability cost, which should be tuned according to the resolution of the traversability map.

The corresponding modified A* algorithm for path planning is presented as in Algorithm. \ref{alg2}. The algorithm terminates when the target point is reached, or there are no paths eligible.

\begin{algorithm} 
\caption{A* path planning on the WTG}\label{alg2}
\textbf{Input:} WTG, start voxel, target voxel \\
\textbf{Output:} Solution path from the start voxel to the target voxel	
\begin{algorithmic}
\State Initialize the \textbf{OPEN} and \textbf{CLOSED} list;
\State Put the start voxel to the \textbf{OPEN} list, set the travel cost of the node as zero;
\While{the \textbf{OPEN} list $\neq $ \O}
    \State Find the node with least travel cost from the \textbf{OPEN} list, set it as node $\mathbf{Q}$;
    \State Pop $\mathbf{Q}$ off the \textbf{OPEN} list;
    \State Generate $\mathbf{Q}$'s successors by checking corresponding connectivity vector $\mathbf{C}_{i.j}^l$, then set their parents to $\mathbf{Q}$;
    \If{$\mathbf{Q}$.successor = $\emptyset$}
        \State \textbf{Return} "No path found";
    \EndIf
    \For{each successor, $\mathbf{S}$}
        \If{the successor, $\mathbf{S}$ == the target}{\textbf{Return} solution path;}
        \Else
            \State $\mathbf{S}$.$\mathbf{g}$ = $\mathbf{Q}$.$\mathbf{g}$ + distance between the successor and $\mathbf{Q}$;
            \State $\mathbf{S}$.$\mathbf{h} =$ distance from the successor to the target voxel;
            \State $\mathbf{S}.\mathbf{f} = \mathbf{S}.\mathbf{g} + \mathbf{S}.\mathbf{h} + \mathbf{S}.\mathbf{F}$ (the element in current $\mathbf{F}_{i,j}^{l}$ in the direction from $\mathbf{Q}$ to the successor $\mathbf{S}$);
            \If{the successor is already in the $\mathbf{OPEN} /\ \mathbf{CLOSED}$ list}
                \If{the successor's travel cost is higher than the existing node}
                    Skip it;
                \Else
                    Replace the existing node;
                \EndIf
            \Else
                \State Add the successor to the \textbf{OPEN} list;
            \EndIf
        \EndIf
    \EndFor
    \State Push $\mathbf{Q}$ on the \textbf{CLOSED} list;
\EndWhile
\end{algorithmic}
\end{algorithm}

\section{Experimental results and discussions}
We provide various path planning cases in this section to demonstrate the effectiveness of the method above. The first group is path planning with a customized UGV system, and the point cloud data are obtained with onboard sensors. The second group is to verify the method in more general environments by using larger maps from online databases. A commercial vehicle is used for such tests. At last, we provide particular case studies to compare the proposed method with the cutting-edge work of 3D path planning in \cite{krusi2017driving, wang2023towards, atas2022elevation}.

\subsection{Experimental tests on a UGV system}
\begin{figure}[h]
    \centering
    \includegraphics[height=2.10in]{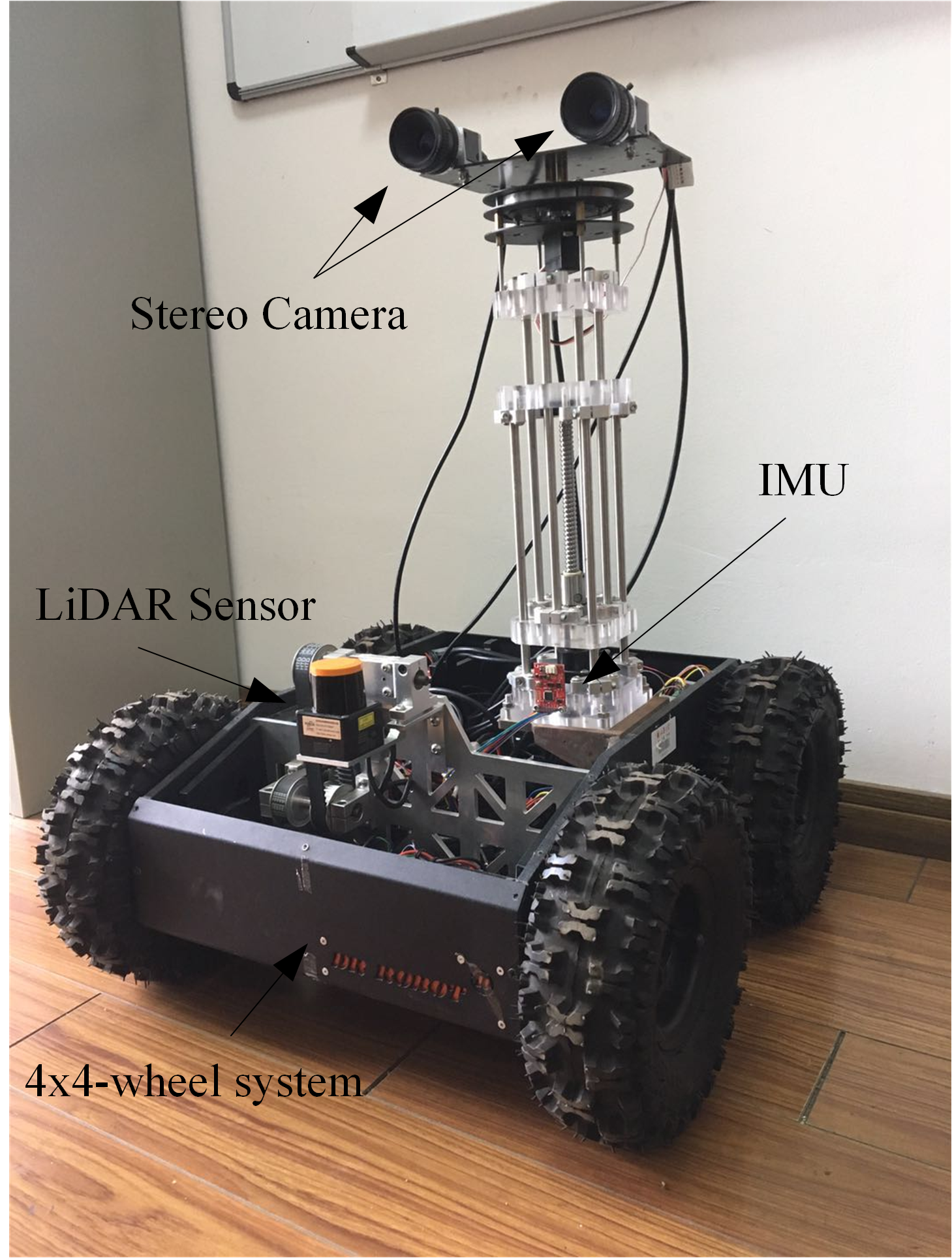}
    \caption{The experimental UGV platform.}
    \label{fig11}
\end{figure}

The algorithms in this article are testified on a UGV platform depicted in Figure \ref{fig11}. The Dc. Robot Jaguar 4x4-wheel chassis system is suitable for both indoor and outdoor operations. Extra sensors and auto-control modules are customized and installed on the system for the purpose of autonomous driving and terrain measurement.

The sensing system is composed of a stereo camera, an IMU (inertial measurement unit), the wheel encoders, and a 3D LiDAR sensor. The first three sensors are mainly used for self-localization when the vehicle is exploring unknown environments. The LiDAR sensor is used to acquire the environmental geometric information. The sensing system provides the vehicle with the capability of dense SLAM.  

\begin{figure}[h]
    \centering
    \includegraphics[height=2.10in]{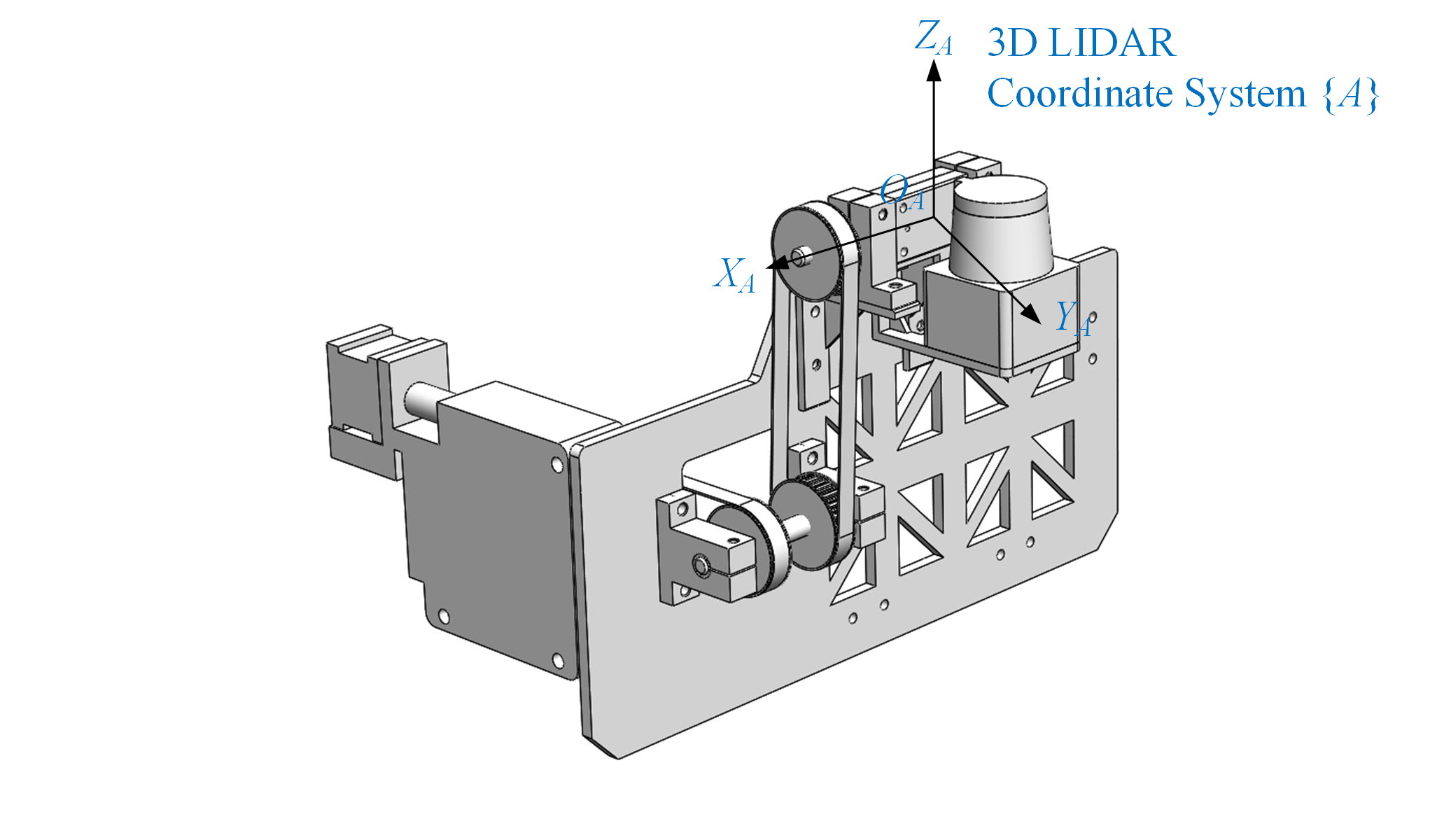}
    \caption{The 3D LiDAR system.}
    \label{fig12}
\end{figure}

The customized dense scanning system is shown in Figure \ref{fig12}. It is mounted on the front of the vehicle and is designed by adding an extra nodding degree of freedom to a 2D LiDAR (HOKUYO UST-10LX), as shown in Figure \ref{fig12}. The average 3D measurement error is 4.84 mm with a detection range of 5m. The angle resolutions of the scanning in the horizontal and vertical directions are both 0.25°. 

Typical indoor and outdoor environments were selected as the testing ground for the path planning algorithm. We first drove the robot around and obtained the point cloud of the test scene using techniques in \cite{xie2020map}. With the point cloud maps, the path planning algorithm in this article can be verified.

\begin{figure}[H]
\centering
     \begin{subfigure}[t]{0.45\textwidth}
         \centering
         \includegraphics[width=\textwidth]{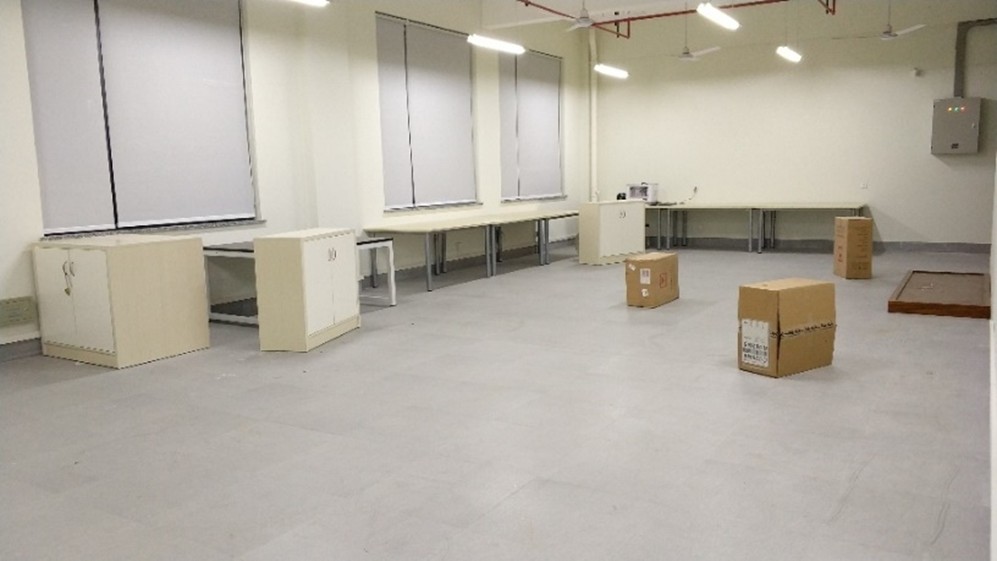}
         \caption{}
         \label{fig:fig13-a}
     \end{subfigure}
     \\
     \begin{subfigure}[t]{0.45\textwidth}
         \centering
         \includegraphics[width=\textwidth]{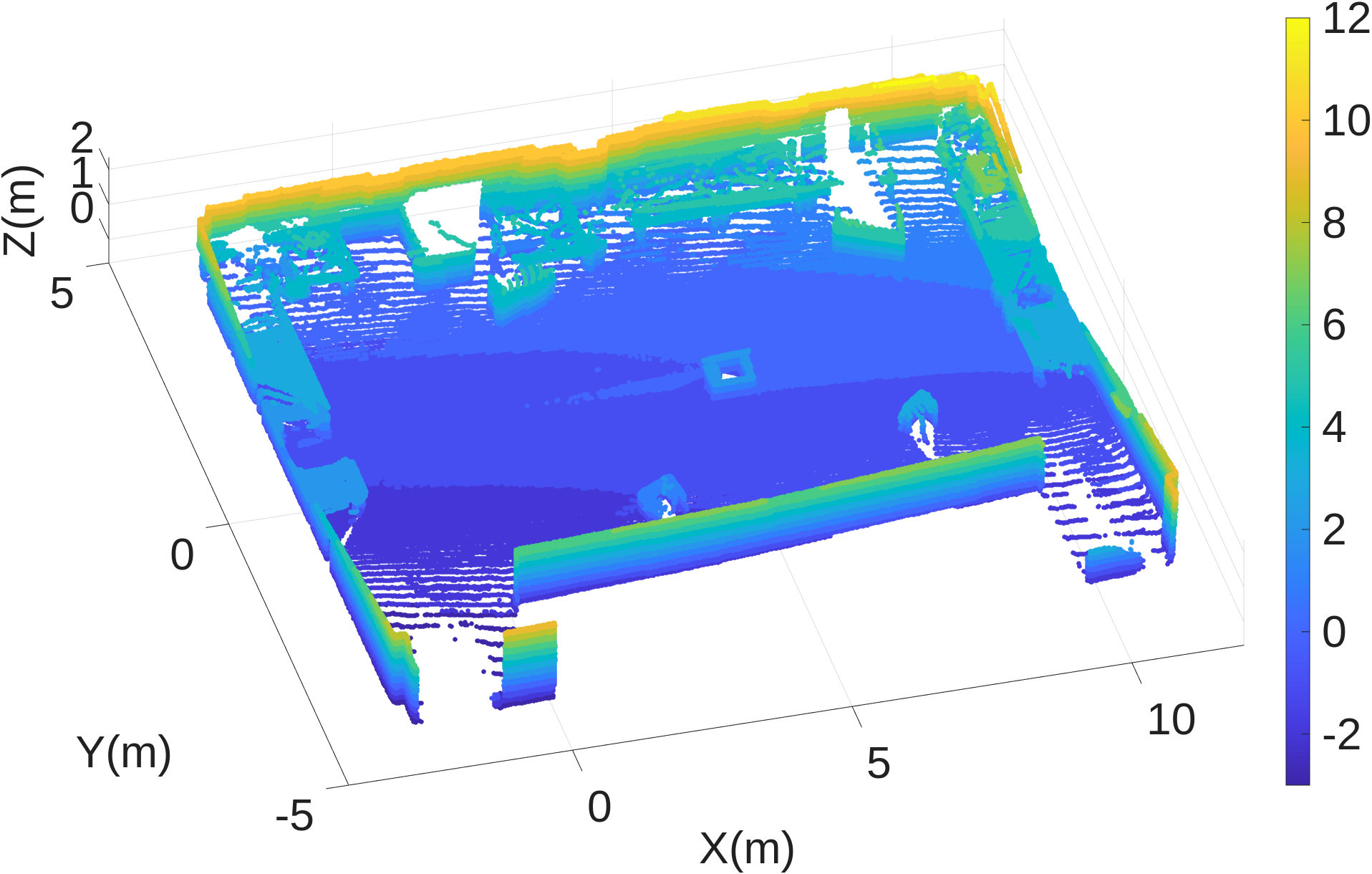}
         \caption{}
         \label{fig:fig13-b}
     \end{subfigure}
     \begin{subfigure}[t]{0.45\textwidth}
         \centering
         \includegraphics[width=\textwidth]{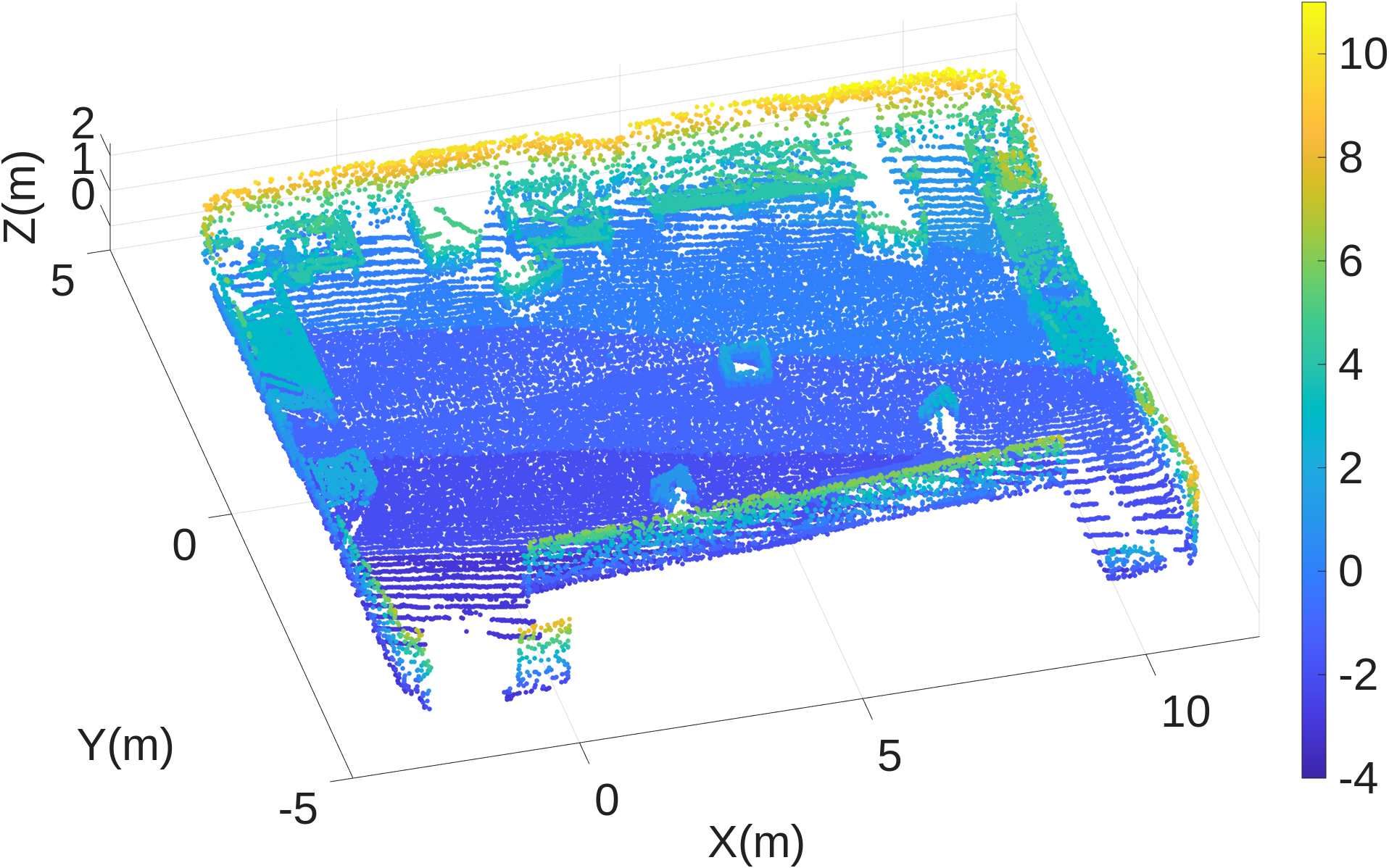}
         \caption{}
         \label{fig:fig13-c}
     \end{subfigure}
     \caption{Indoor environment mapping results. (a) Indoor test area with boxes and furniture serving as obstacles. (b) Dense point cloud acquired using a 3D LiDAR sensor. (c) Variable-resolution map generated by the proposed method. } \label{fig13}
\end{figure}

Figure \ref{fig13}(a) shows an indoor environment with boxes and furniture placed around as obstacles. With the 3D LiDAR scanner, the point clouds are obtained as in Figure \ref{fig17}(b). The original map has 937260 points, the number of which is decreased to be 88668 as a variable resolution map in Figure \ref{fig17}(c). The redundant points of the flat floor and the wall were removed, while the points of the edges and corners of the obstacles are well preserved.

\begin{figure}[H]
\centering
     \begin{subfigure}[t]{0.45\textwidth}
         \centering
         \includegraphics[width=\textwidth]{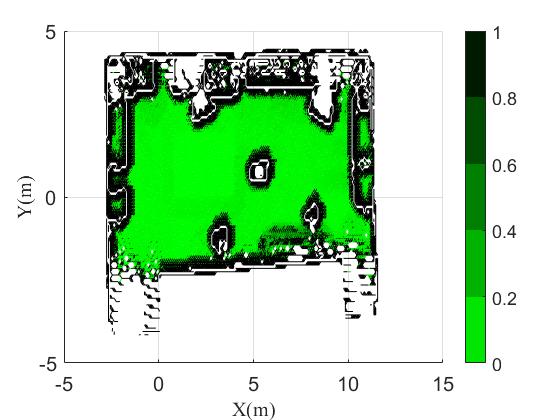}
         \caption{}
         \label{fig:fig14-a}
     \end{subfigure}
     \begin{subfigure}[t]{0.45\textwidth}
         \centering
         \includegraphics[width=\textwidth]{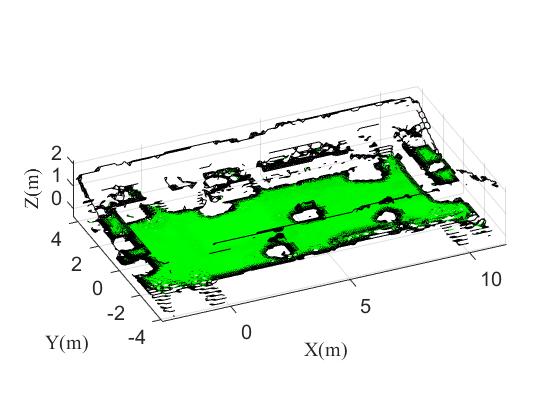}
         \caption{}
         \label{fig:fig14-b}
     \end{subfigure}
     \caption{The traversability map of the indoor scene from: (a) the top view and (b) the side view. The green regions are traversable, and the black regions are unpassable.  } \label{fig14}
\end{figure}

With the traversability analysis method in Section \ref{sec:Weighted_traversability_graph}, the WTG is obtained and shown in Figure \ref{fig14} with two different viewpoints. Most of the terrain is denoted with a low traversability cost, and the walls and furniture are directly classified as untraversable places. The ground is a traversable surface, and the up surfaces of the furniture and walls are in different levels with the ground. Though small areas on the tables are recognized as local traversable, there is no possibility to travel from the table to the ground since their surface voxels are not connected. 

\begin{figure}[H]
\centering
     \begin{subfigure}[t]{0.45\textwidth}
         \centering
         \includegraphics[width=\textwidth]{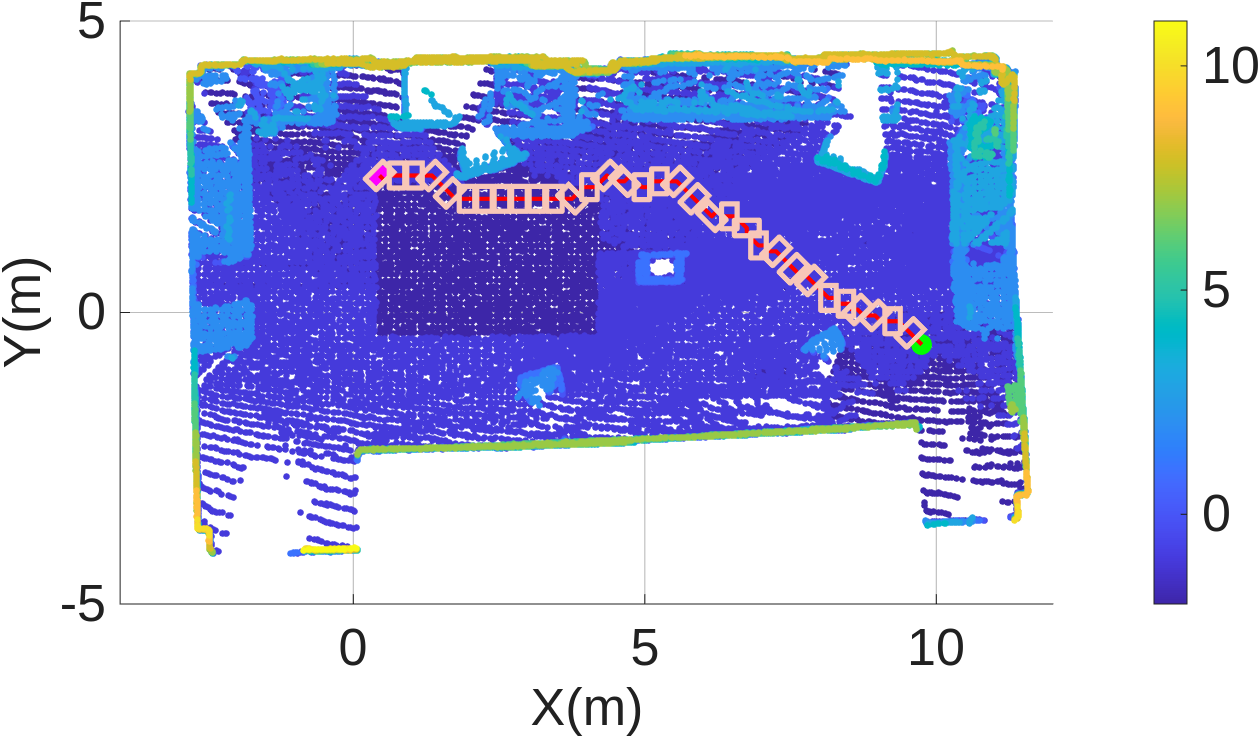}
         \caption{}
         \label{fig:fig15-a}
     \end{subfigure}
     \begin{subfigure}[t]{0.45\textwidth}
         \centering
         \includegraphics[width=\textwidth]{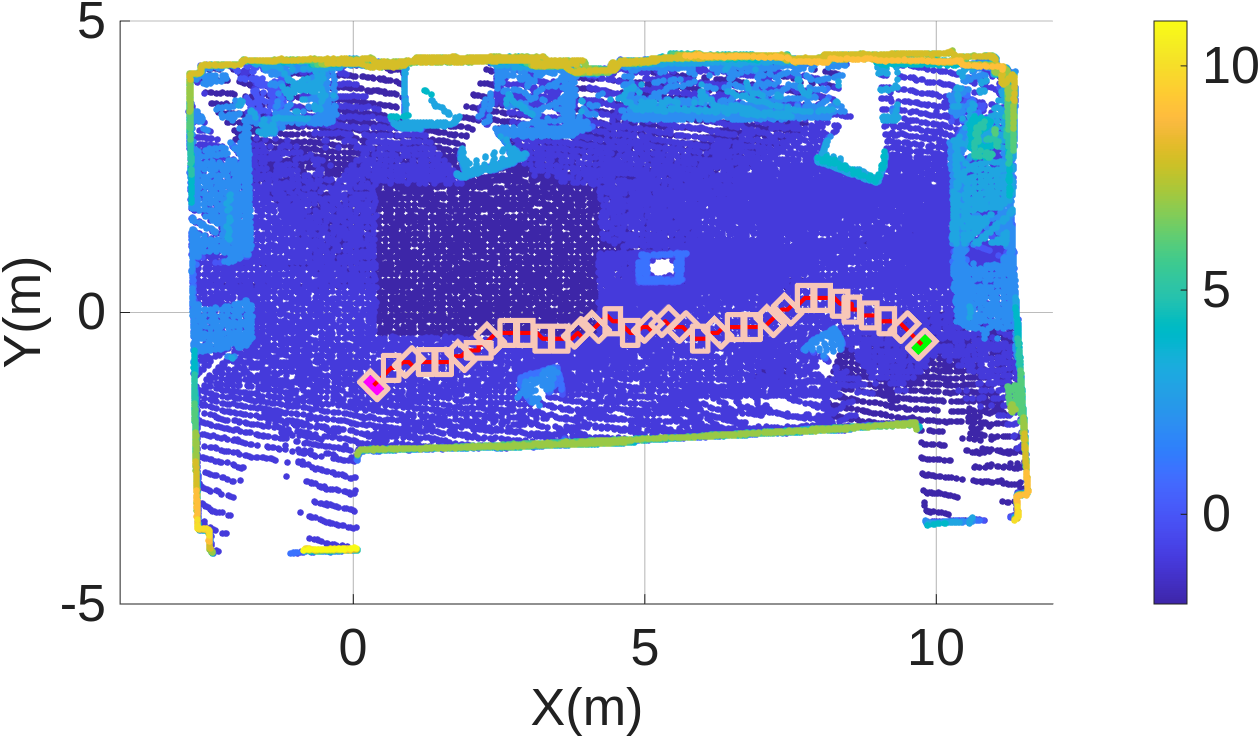}
         \caption{}
         \label{fig:fig15-b}
     \end{subfigure}
     \caption{Indoor path planning for two pairs of start and target points. The red lines are the planned path of the vehicle center, and the square shapes are the corresponding vehicle projection to the $xy$ planes. } \label{fig15}
\end{figure}

Following the path planning method in Section \ref{sec:Graph-based_path_search}, path planning results for two pairs of starting and ending points are shown in Figure \ref{fig15}. The planned path and the corresponding projection of the vehicle bodies on the x-y plane indicate that the vehicle can automatically avoid the obstacles on the ground and find a reasonably short path between arbitrary pairs of points.

\begin{figure}[H]
\centering
     \begin{subfigure}[t]{0.45\textwidth}
         \centering
         \includegraphics[width=\textwidth]{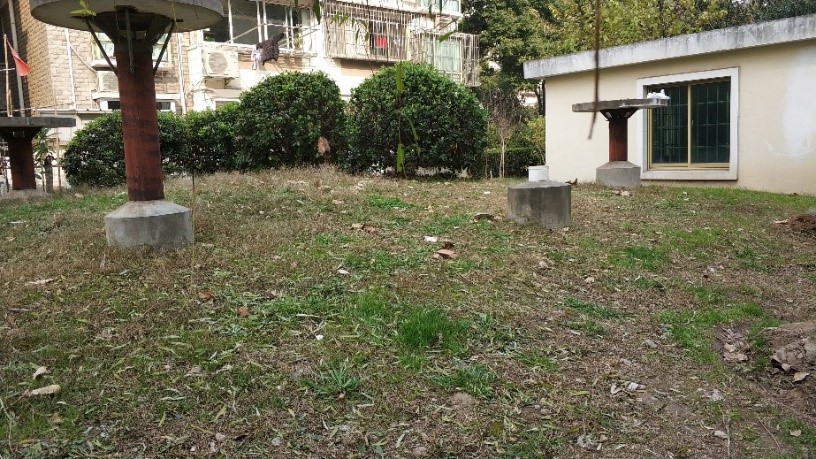}
         \caption{}
         \label{fig:fig16-a}
     \end{subfigure}
     \\
     \begin{subfigure}[t]{0.45\textwidth}
         \centering
         \includegraphics[width=\textwidth]{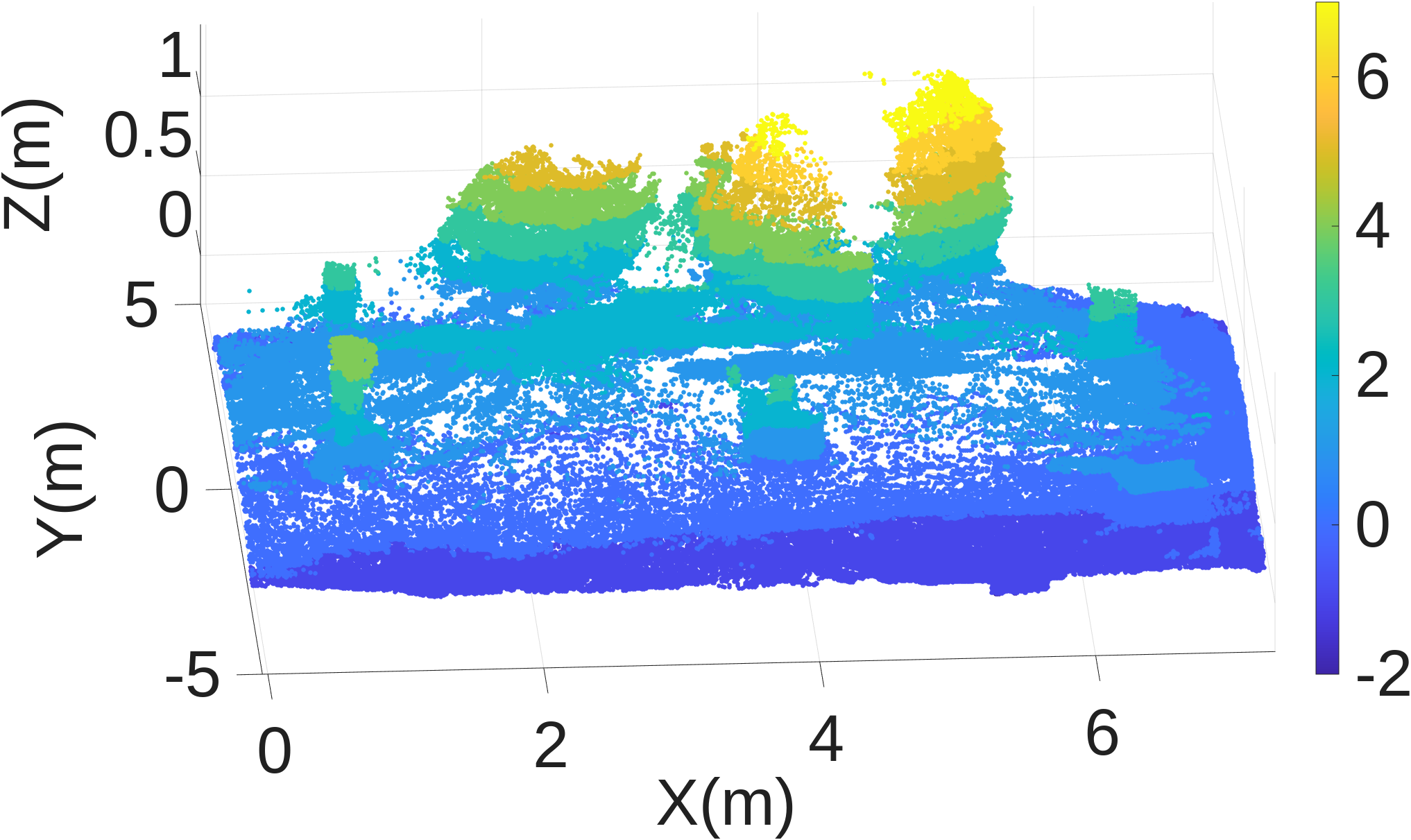}
         \caption{}
         \label{fig:fig16-b}
     \end{subfigure}
     \begin{subfigure}[t]{0.45\textwidth}
         \centering
         \includegraphics[width=\textwidth]{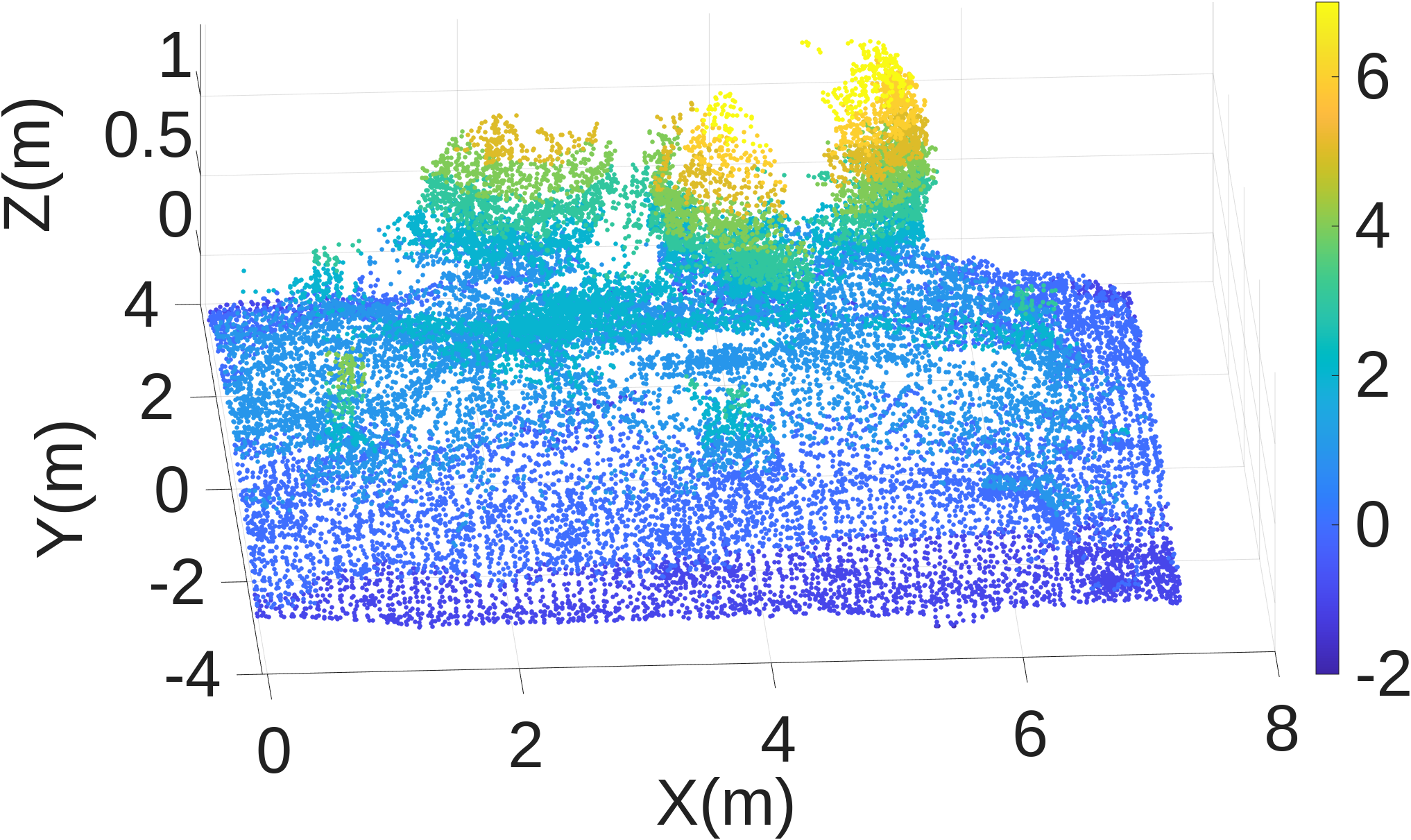}
         \caption{}
         \label{fig:fig16-c}
     \end{subfigure}
     \caption{ The first test scene. (a) The picture of the motto. (b) Its dense point cloud map. (c) Its variable resolution map.  } \label{fig16}
\end{figure}

\begin{figure}[H]
\centering
     \begin{subfigure}[t]{0.45\textwidth}
         \centering
         \includegraphics[width=\textwidth]{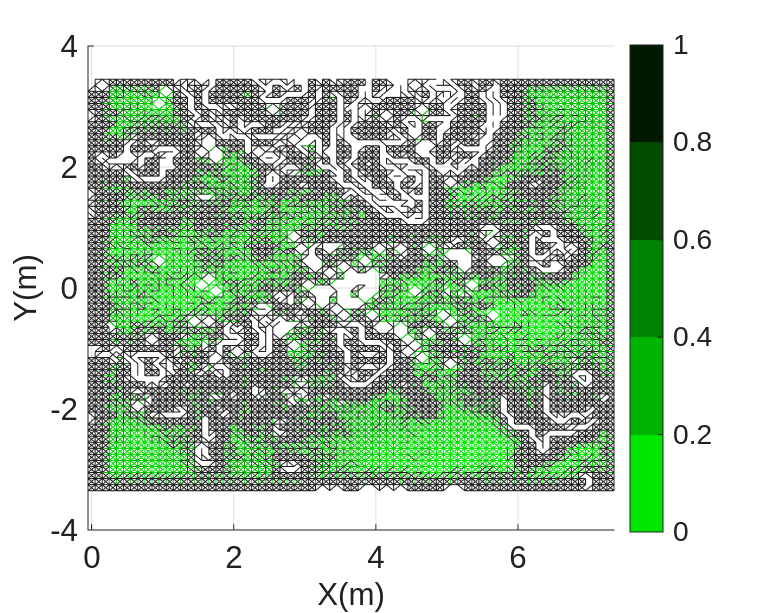}
         \caption{}
         \label{fig:fig17-a}
     \end{subfigure}
     \begin{subfigure}[t]{0.45\textwidth}
         \centering
         \includegraphics[width=\textwidth]{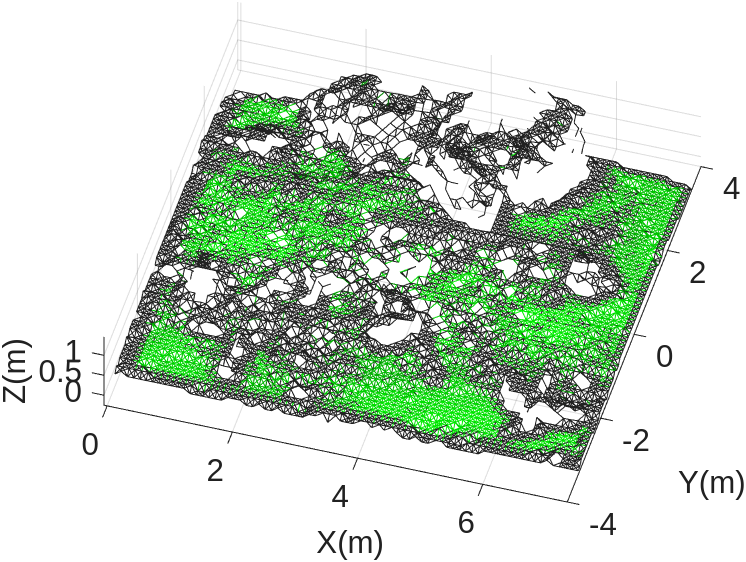}
         \caption{}
         \label{fig:fig17-b}
     \end{subfigure}
    \caption{The traversability map of the motto. (a) From the top view. (b) From side view. }
    \label{fig17}
\end{figure}

\begin{figure}[H]
\centering
     \begin{subfigure}[t]{0.45\textwidth}
         \centering
         \includegraphics[width=\textwidth]{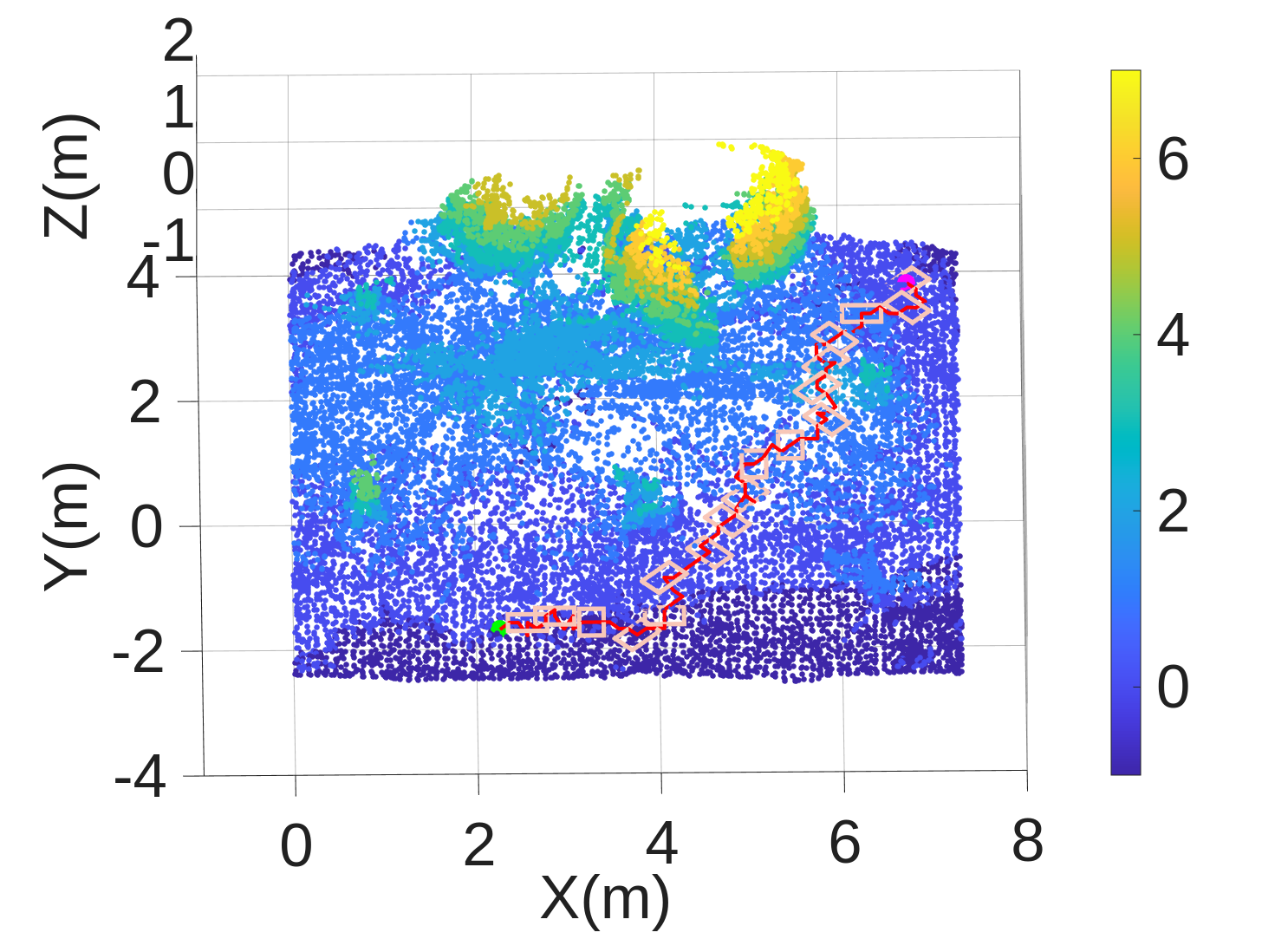}
         \caption{}
         \label{fig:fig18-a}
     \end{subfigure}
     \begin{subfigure}[t]{0.45\textwidth}
         \centering
         \includegraphics[width=\textwidth]{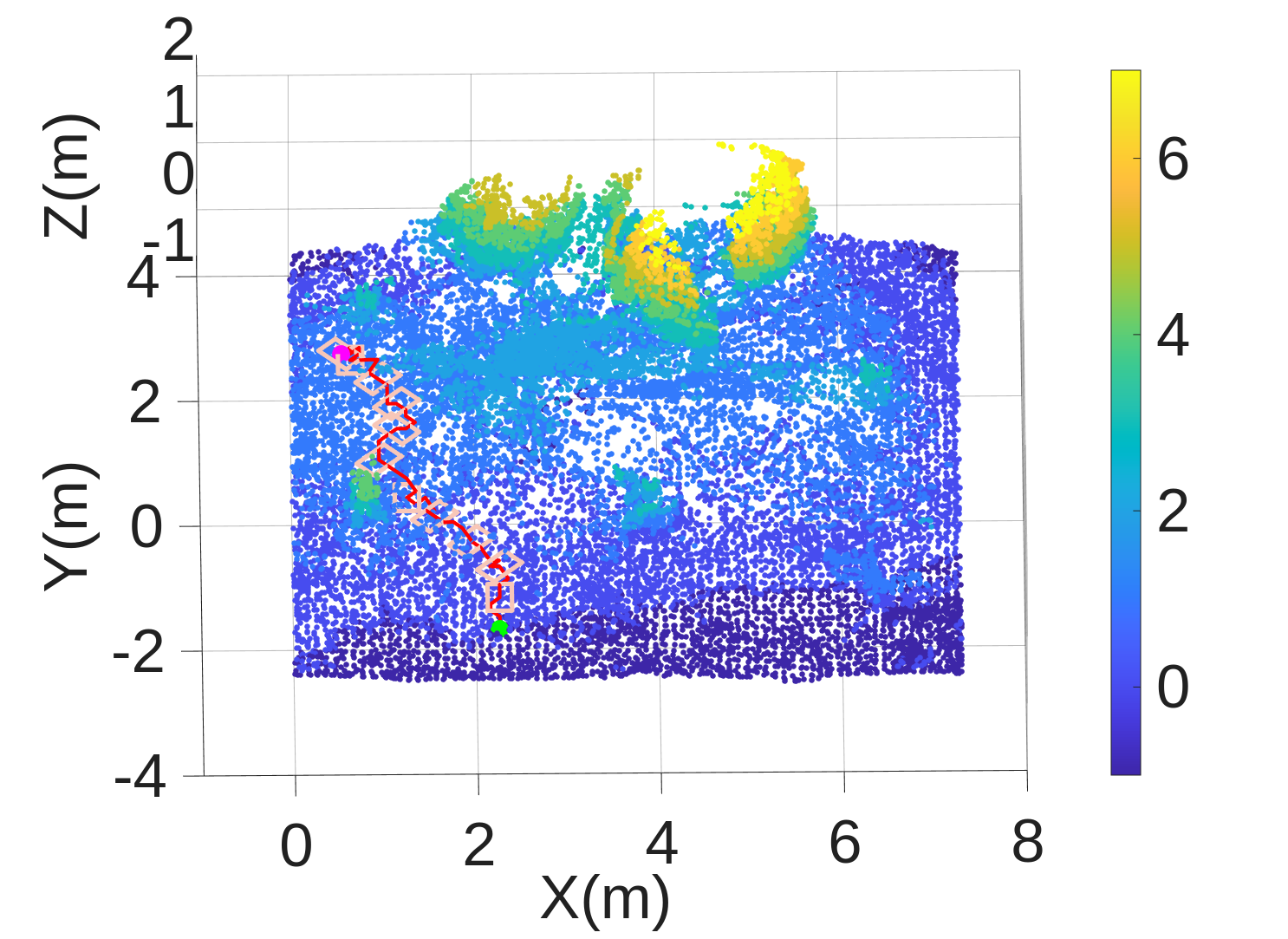}
         \caption{}
         \label{fig:fig18-b}
     \end{subfigure}
    \caption{Path planning for two pairs of start and target points in the motto scene. The red lines are the planned path of the vehicle center, and the square shapes are the corresponding vehicle projection to the planes.}
    \label{fig18}
\end{figure}

Two outdoor terrains were tested. The first one shown in Figure \ref{fig16}(a) is a motte with a relatively small slope angle but a rough terrain surface and various obstacles. Due to the irregularity of the terrain surface, the point cloud generated by laser scanners has several occlusion areas, as shown in Figure \ref{fig16}(b) and (c). The original measurement has 334803 points, which is down to 70257 in the variable resolution map. The WTG in Figure \ref{fig17} shows that the traversability condition of this terrain is much complicated, mainly due to the missing pieces of the terrain surface, the chassis impact hazards, and the overhanging plants. Even under such complex conditions, the algorithm successfully finds safe ways for the vehicle to pass with relatively small slopes and without collision with the ground or the obstacles, as shown in Figure \ref{fig18}. 

\begin{figure}[H]
\centering
     \begin{subfigure}[t]{0.45\textwidth}
         \centering
         \includegraphics[width=\textwidth]{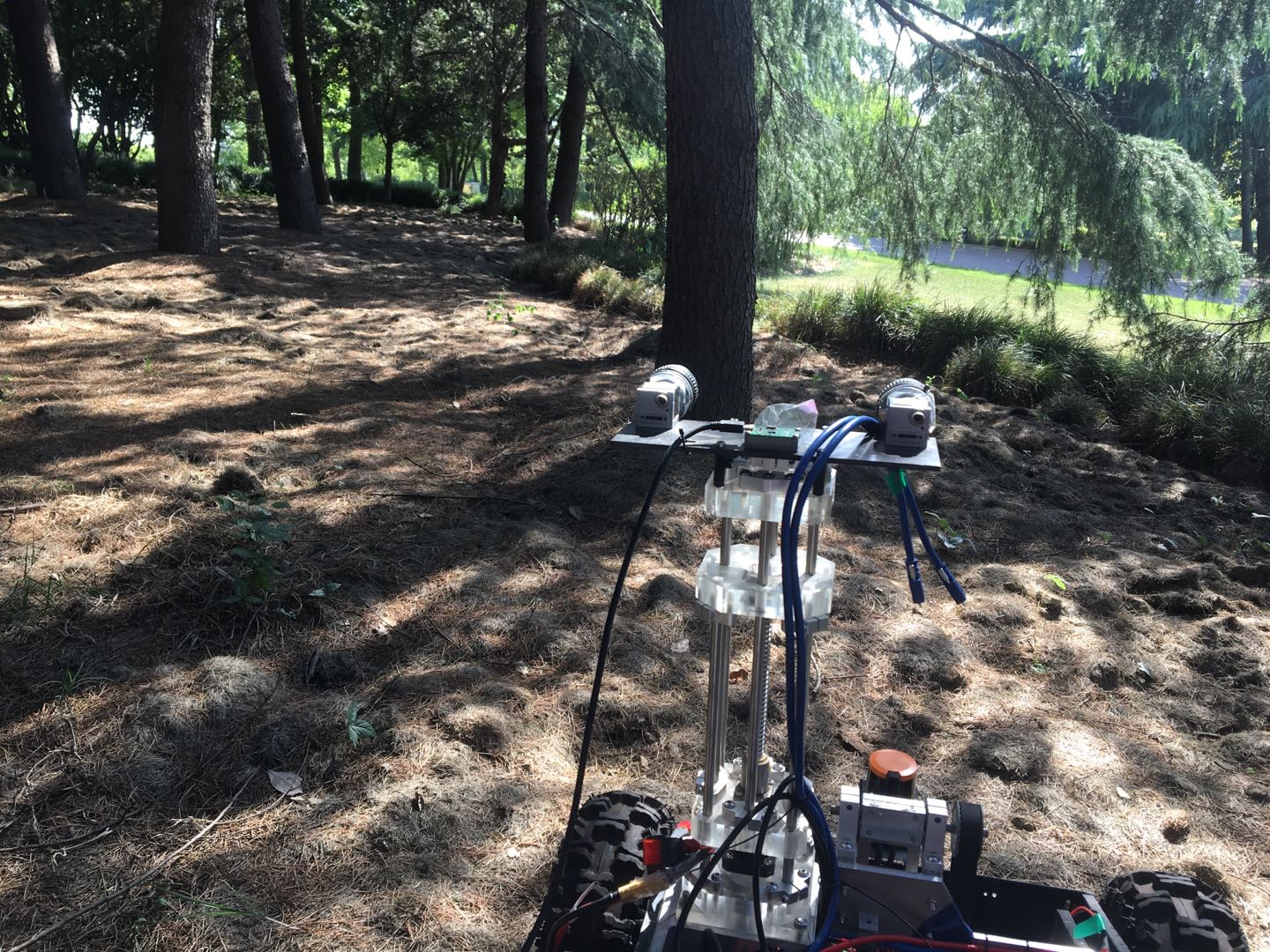}
         \caption{}
         \label{fig:fig19-a}
     \end{subfigure}
     \begin{subfigure}[t]{0.45\textwidth}
         \centering
         \includegraphics[width=\textwidth]{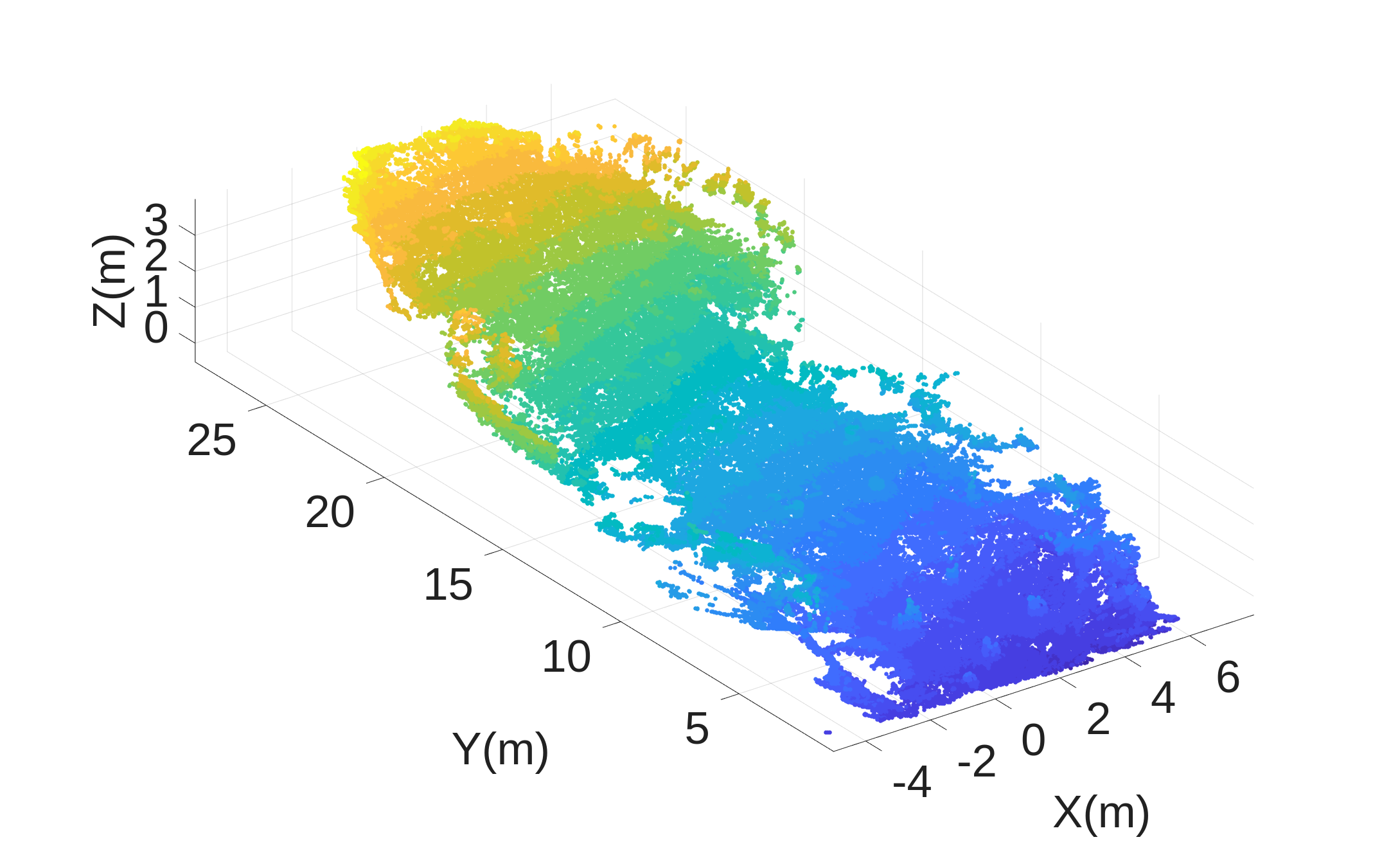}
         \caption{}
         \label{fig:fig19-b}
     \end{subfigure}
    \caption{The second outdoor test scene. (a) The picture of the grove. (b) Its point cloud map. }
    \label{fig19}
\end{figure}

\begin{figure} [H]
    \centering
    \includegraphics[height=2.50in]{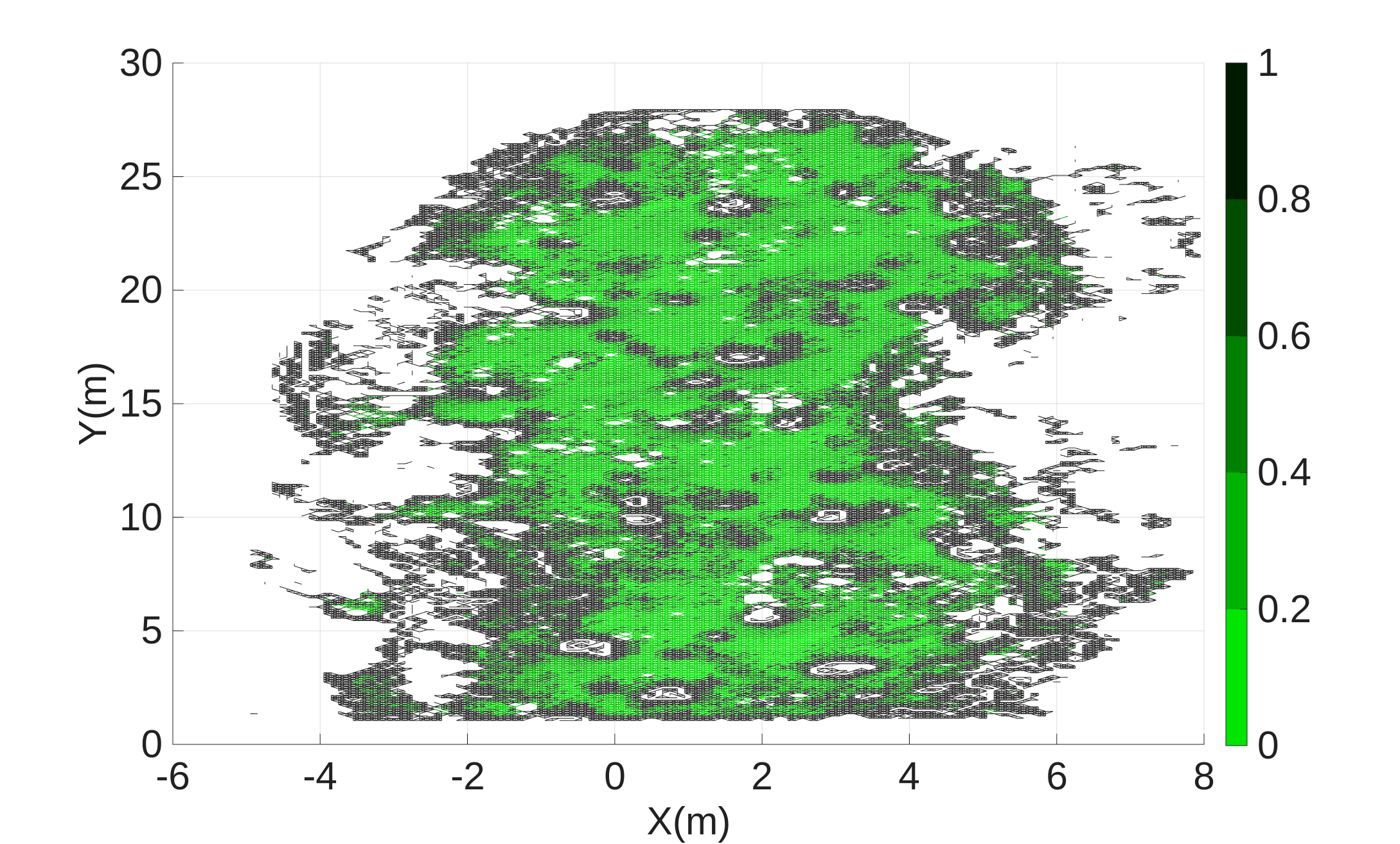}
    \hspace{.2in}
    \caption{The traversability map of the grove from the top view.}
    \label{fig20}
\end{figure}

\begin{figure} [H]
    \centering
    \includegraphics[height=2.30in]{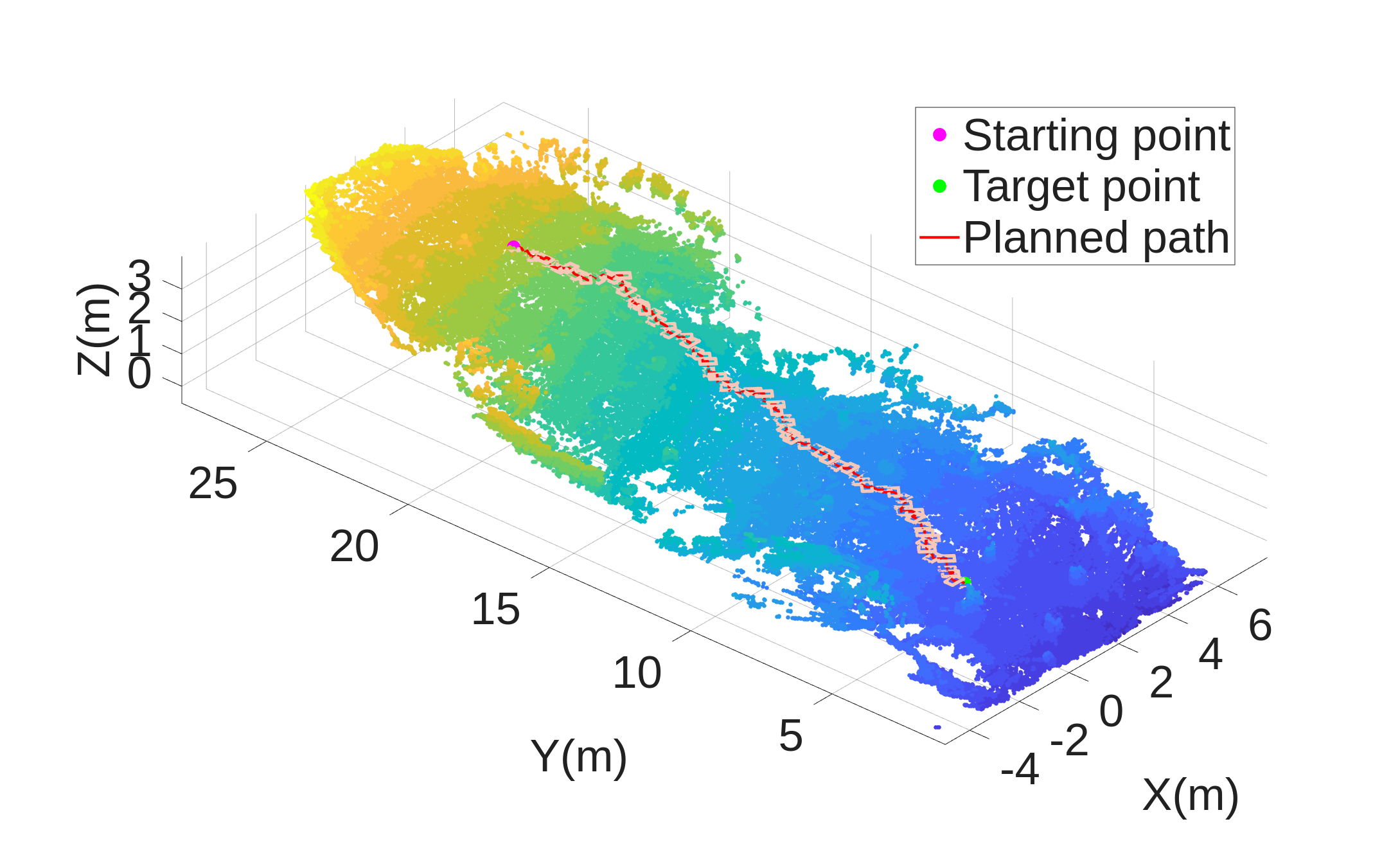}
    \hspace{.2in}
    \caption{Path planning results in the grove scene. The red lines are the planned path of the vehicle center, and the square shapes are the corresponding vehicle projection to the $xy$ planes. }
    \label{fig21}
\end{figure}

The second outdoor test scene is a grove with pine trees growing close together and a relatively uneven terrain underneath. Figure \ref{fig19} shows the test scene from the robot's perspective. The ground is irregular with small bumps and slopes, and there are dozens of trees that the robot should keep away from. The WTG is shown in Figure \ref{fig20}. The areas with bushes and trees are denoted as untraversable. Noticeably, because the terrain is a continuous plane without a large height difference, there is rarely a discontinuous area on the ground. Given two points as the start and target points at the two ends of the grove, the planned path from the proposed planner is shown in Figure \ref{fig21}. The path is close to a straight line connecting the two points but carefully avoids the collisions with trees on the way.

To evaluate the robustness of the proposed method, we conduct systematic experiments 
under varying noise conditions and point cloud sparsity levels using the grove 
scenario. Gaussian noise is added to each LiDAR point to simulate random measurement 
disturbances, with noise levels controlled by standard deviation $\sigma$ ranging 
from 0.005 m to 0.06 m. These values correspond to different operational conditions: 
$\sigma = 0.005$ m represents high-quality close-range indoor measurements, 
$\sigma = 0.02$ m represents standard outdoor conditions, $\sigma = 0.035$ m simulates 
adverse weather or vegetation interference, and $\sigma = 0.06$ m models extreme-range 
measurements in heavy rain or fog. Additionally, point cloud sparsity is tested by 
uniformly subsampling the original data at rates from 90\% to 30\% density.

Five target positions are selected within the grove scene to evaluate planning 
performance. Success rate is defined as the percentage of targets for which the 
planner successfully finds a valid path. Results are presented in Table~\ref{table2}. 
The proposed method demonstrates robustness to moderate noise levels ($\sigma \leq 0.035$ m), 
as the ML-SkiMap's variable-resolution representation effectively filters measurement 
noise that does not significantly alter terrain slope or geometric features critical 
for traversability assessment. At higher noise levels ($\sigma = 0.06$ m), noise 
effectively inflates obstacle boundaries, potentially blocking narrow passages; 
however, in the grove scenario with wide inter-tree spacing, this inflation does not 
affect the success rate. Similarly, point cloud subsampling at moderate rates 
(down to 50\% density) does not degrade planning performance, indicating sufficient 
geometric information is preserved. However, at 30\% density, extensive occlusion 
regions trigger the conservative traversability policy described in 
Section~\ref{sec_occlusion}, causing planning failures for most of the targets as the planner 
avoids traversing unknown areas. In summary, the proposed method maintains robust 
performance with noise levels up to $\sigma = 0.035$ m and point cloud densities 
down to 30\%, exhibiting conservative but safe behavior under more severe degradation. 

\begin{table}[H]
\centering
\begin{tabular}{|l|l|l|l|l|l}
\cline{1-5}
\textbf{Gaussian noise deviation (m)} & \textbf{0.01} & \textbf{0.02} & \textbf{0.035} & \textbf{0.06} &  \\ \cline{1-5}
Success rate & 100\%         & 100\%         & 100\%          & 100\%         &  \\ \cline{1-5}
\textbf{Subsample rate}               & \textbf{90\%} & \textbf{70\%} & \textbf{50\%}  & \textbf{30\%} &  \\ \cline{1-5}
Success rate & 100\%         & 100\%         & 100\%          & 20\%          &  \\ \cline{1-5}
\end{tabular}
\caption{Success rate of the proposed path planning method under different measurement conditions.} \label{table2}
\end{table}

\subsection{Case study with online database}\label{sec:case_study_online_database}
\begin{figure} [H]
    \centering
    \includegraphics[height=1.50in]{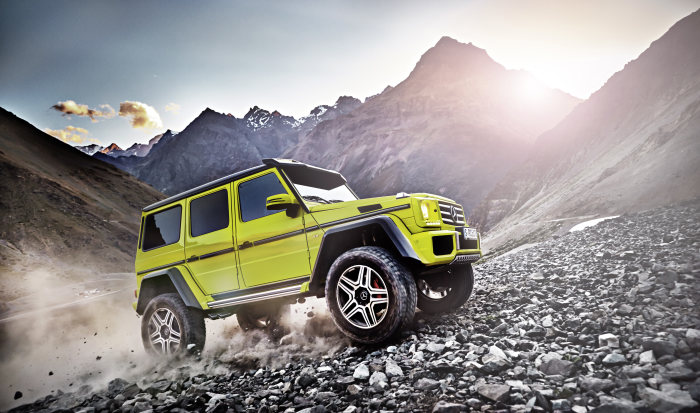}
    \hspace{.2in}
    \caption{Off-road vehicle, Mercedes-Benz G 500 4×4².}
    \label{fig22}
\end{figure}

In this section, the testing grounds are three more complex scenarios obtained from online resources. They include not only the rough and wild terrains but also multi-floor urban architectures. A larger vehicle, Mercedes-Benz G500, as in Figure \ref{fig22}, is used for the tests. The relevant vehicle parameters are shown in table \ref{table1}.

\begin{table}[H]
\resizebox{\linewidth}{!}{
\begin{tabular}{|l|l|l|l|}
\hline
Category & Description & Symbol & Value\\ 
\hline
\multirow{3}{*}{Mapping} & Voxel size & $d\times d \times d_z$ & $500\times 500\times 500$ mm \\
\cline{2-4} 
                  & Variant resolution map parameter                   & {[}a,b,c{]}         & {[}900,3,1{]}  \\ \cline{2-4} 
                  & Minimum distance between two layers                & $h_L$               & 3000 mm        \\ \hline
\multirow{4}{*}{Vehicle dimensions}        & Half of the distance between left and right wheels & W                   & 887 mm         \\ \cline{2-4} 
                  & Half of the distance between front and rear wheels & L                   & 1425 mm        \\ \cline{2-4} 
                  & The radius of the wheel   & R                   & 458.15 mm      \\ \cline{2-4} 
                  & Ground clearance of the vehicle                    & H                   & 3000 mm        \\ \hline
\multirow{2}{*}{Traversability properties} & Largest allowed tilt angle                         & $\alpha_w$ & 30°            \\ \cline{2-4} 
                  & Maximum vertical step height the robot can traverse        & $h_\text{climb}$           & 500 mm         \\ \hline
\end{tabular}
}
 \caption{Path planning related parameters in the tests using Mercedes-Benz G500}\label{table1}
\end{table}

\begin{figure}[H]
\centering
     \begin{subfigure}[t]{0.31\textwidth}
         \centering
         \includegraphics[width=\textwidth]{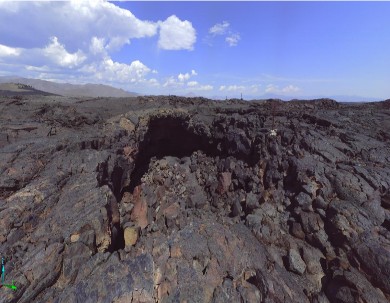}
         \caption{}
         \label{fig:fig23-a}
     \end{subfigure}
     \begin{subfigure}[t]{0.31\textwidth}
         \centering
         \includegraphics[width=\textwidth]{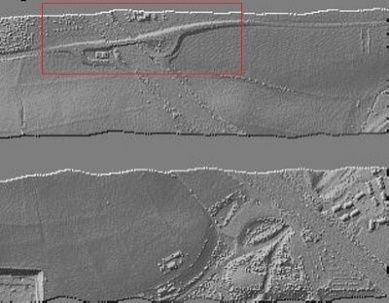}
         \caption{}
         \label{fig:fig23-b}
     \end{subfigure}
     \begin{subfigure}[t]{0.31\textwidth}
         \centering
         \includegraphics[width=\textwidth]{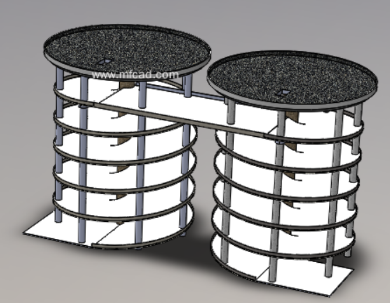}
         \caption{}
         \label{fig:fig23-c}
     \end{subfigure}
    \caption{The original test scenes from the online databases.}
    \label{fig23}
\end{figure}

The first scenario is chosen from the Planetary Pits and Caves 3D Dataset \cite{nasa}, which builds models for many terrestrial pits and caves with high-resolution LIDAR scanning. The IndianTunnel Cave Model in the dataset is used here (Figure \ref{fig23}(a)). There are slopes, large rocks, and semi-open caves on the site. The terrain seems to be quite uneven and could be dangerous to traverse through. The second scenario is a test site in the OEEPE project on laser scanning \cite{sithole2003report}, which is shown in Figure \ref{fig23}(b). Part of the site (Circled in a red box in Figure \ref{fig23}(b)) is chosen to evaluate our path planning algorithm, which includes a typical building feature - the bridge. The last scenario is a multistory parking building, which is generated from an open-source 3D full-scale model, as represented in Figure \ref{fig23}(c).

\begin{figure}[H]
\centering
     \begin{subfigure}[t]{0.45\textwidth}
         \centering
         \includegraphics[width=\textwidth]{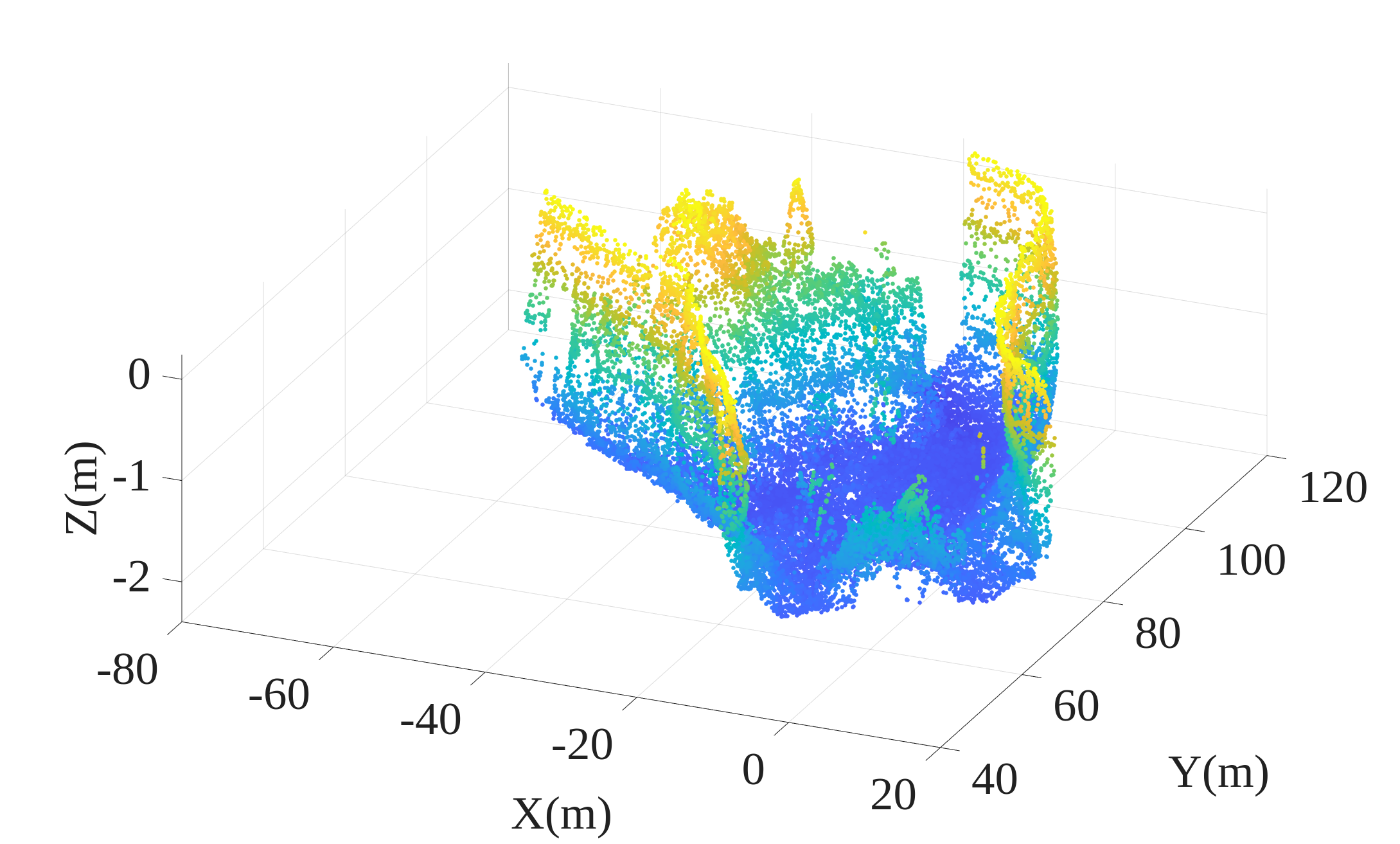}
         \caption{}
         \label{fig:fig24-a}
     \end{subfigure}
     \begin{subfigure}[t]{0.45\textwidth}
         \centering
         \includegraphics[width=0.8\textwidth]{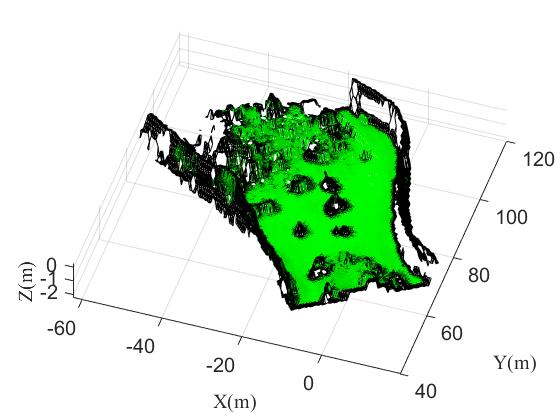}
         \caption{}
         \label{fig:fig24-b}
     \end{subfigure}
     \\
     \begin{subfigure}[t]{0.45\textwidth}
         \centering
         \includegraphics[width=\textwidth]{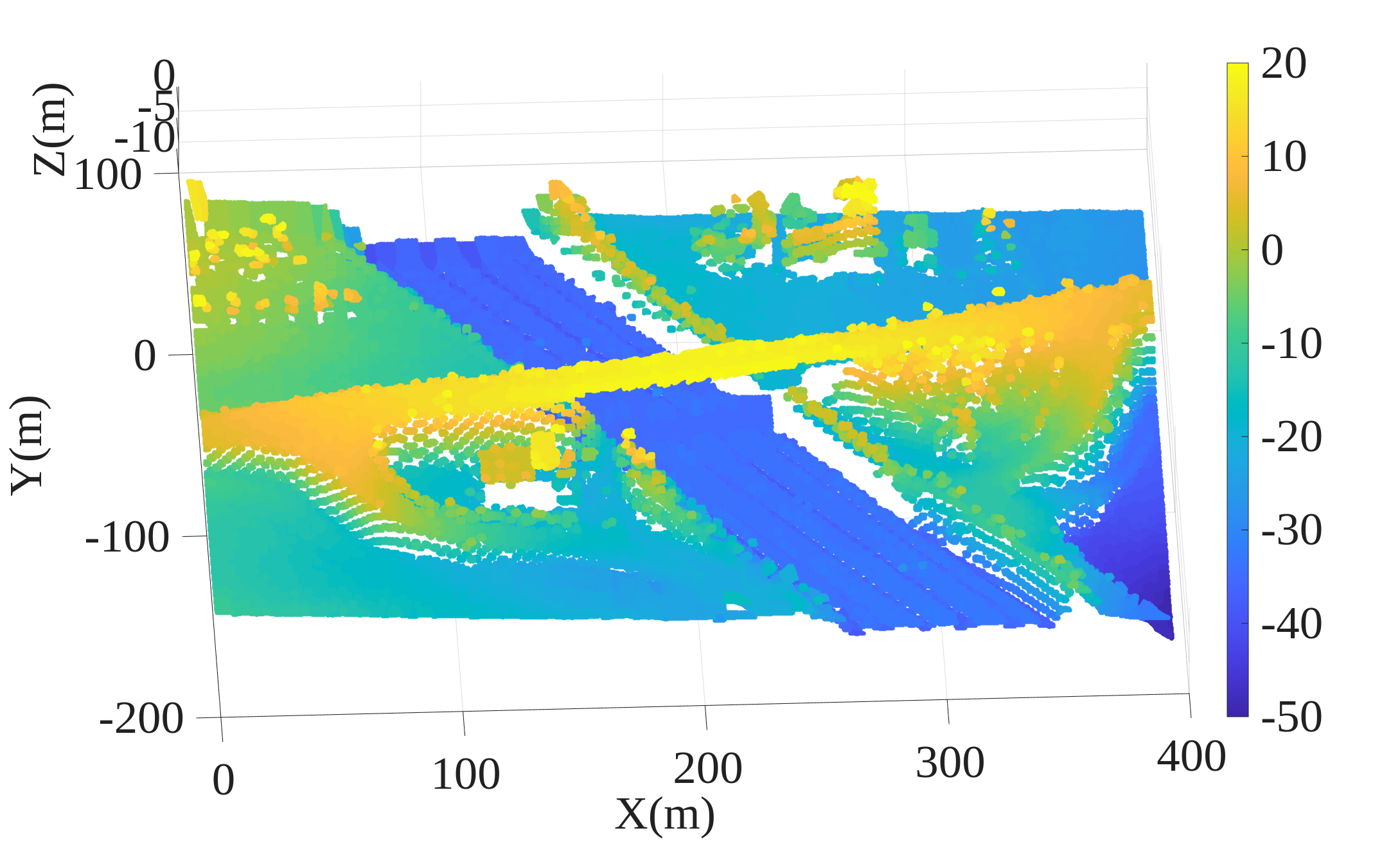}
         \caption{}
         \label{fig:fig24-c}
     \end{subfigure}
     \begin{subfigure}[t]{0.45\textwidth}
         \centering
         \includegraphics[width=0.8\textwidth]{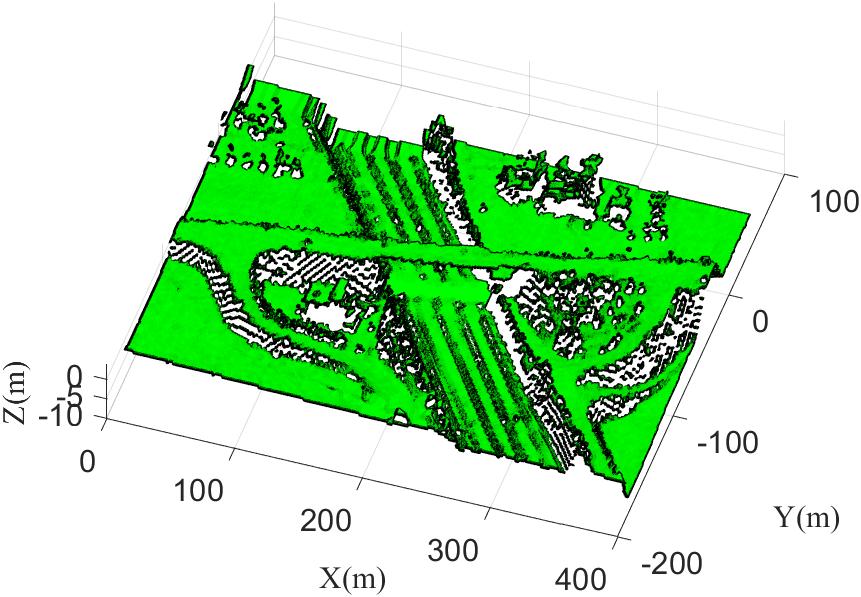}
         \caption{}
         \label{fig:fig24-d}
     \end{subfigure}
     \\
     \begin{subfigure}[t]{0.45\textwidth}
         \centering
         \includegraphics[width=\textwidth]{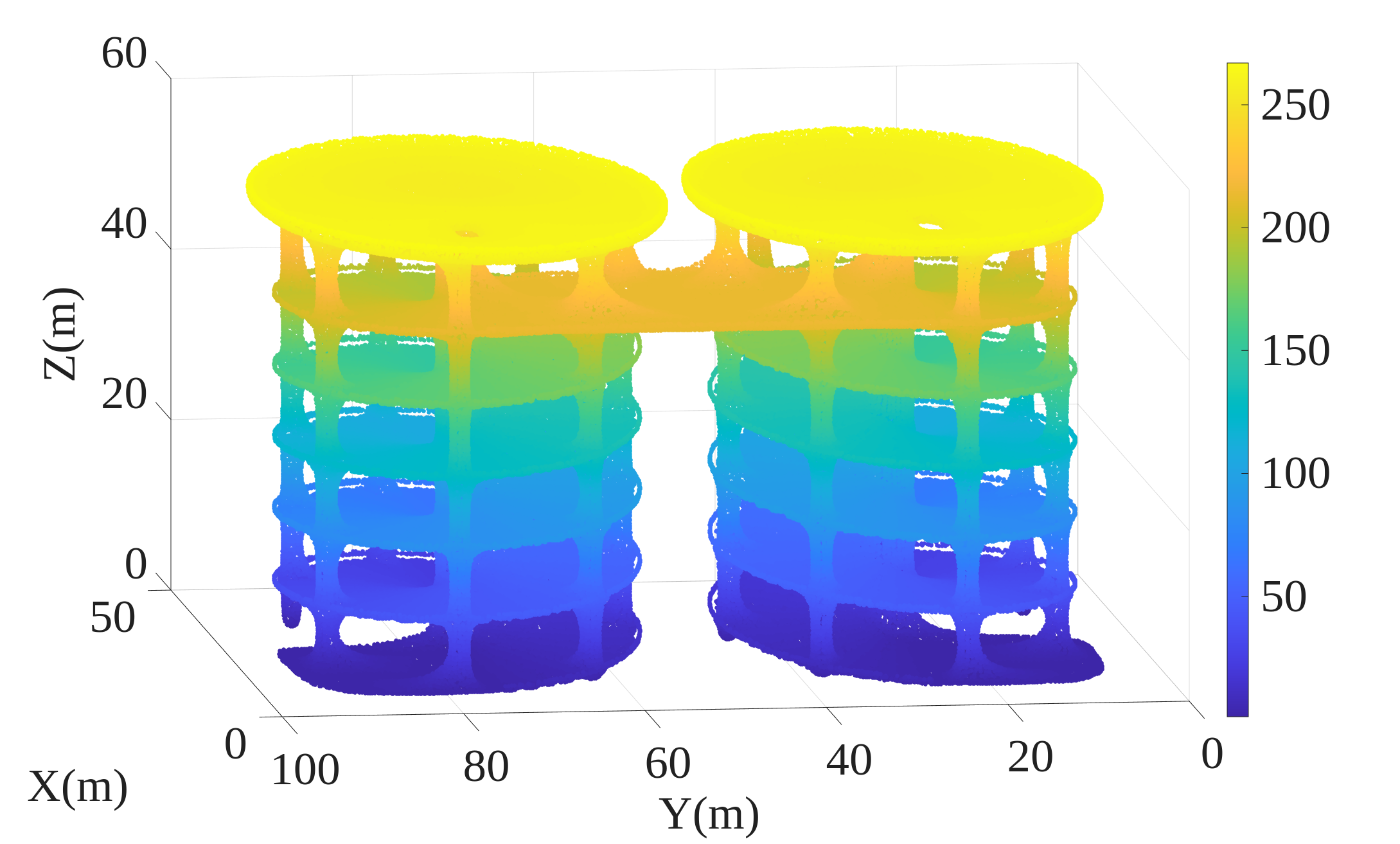}
         \caption{}
         \label{fig:fig24-e}
     \end{subfigure}
     \begin{subfigure}[t]{0.45\textwidth}
         \centering
         \includegraphics[width=0.8\textwidth]{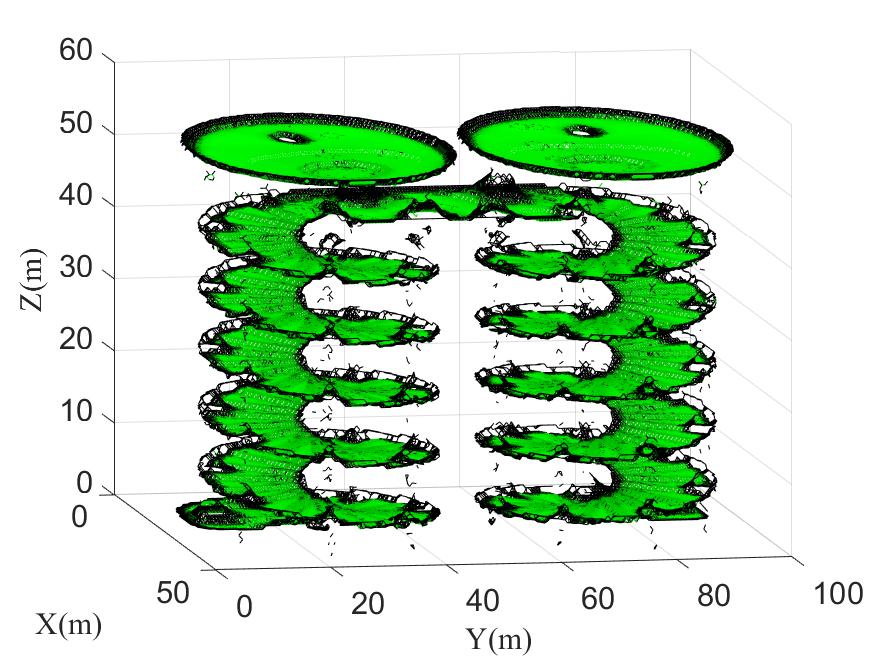}
         \caption{}
         \label{fig:fig24-f}
     \end{subfigure}
    \caption{The variable resolution point clouds of the online datasets (shown as the left pictures) and the corresponding WTGs (shown as the right pictures).}
    \label{fig24}
\end{figure}

The point clouds of the three sites are shown on the left side of Figure \ref{fig24}. Traversability maps of them are obtained using the traversability analysis method proposed in Chapter 3, as shown on the right side of Figure \ref{fig24}. For the three-dimensional structures, the traversability is no longer determined by the height information of the terrain. The point clouds are additionally divided into several layers with regard to their height difference. And the traversability is analyzed for the ground of each layer. For example, in Figure \ref{fig24}(b), the river is clearly on the sub-layer of the bridge, and there is no possible way to pass the river other than the bridge. In Figure \ref{fig24}(c), the connectivity of the spiral ground is clearly shown, and the only passway between the twin buildings on the top floor. The boundary areas of the passway is not passable due to the barriers, and the roofs of the buildings are separated from other part and thus unreacheable. The WTGs accurately extract the geometric structure of the terrain and reflect the local traversability of the ground surfaces. 

\begin{figure} [H]
    \centering
    \includegraphics[height=2.10in]{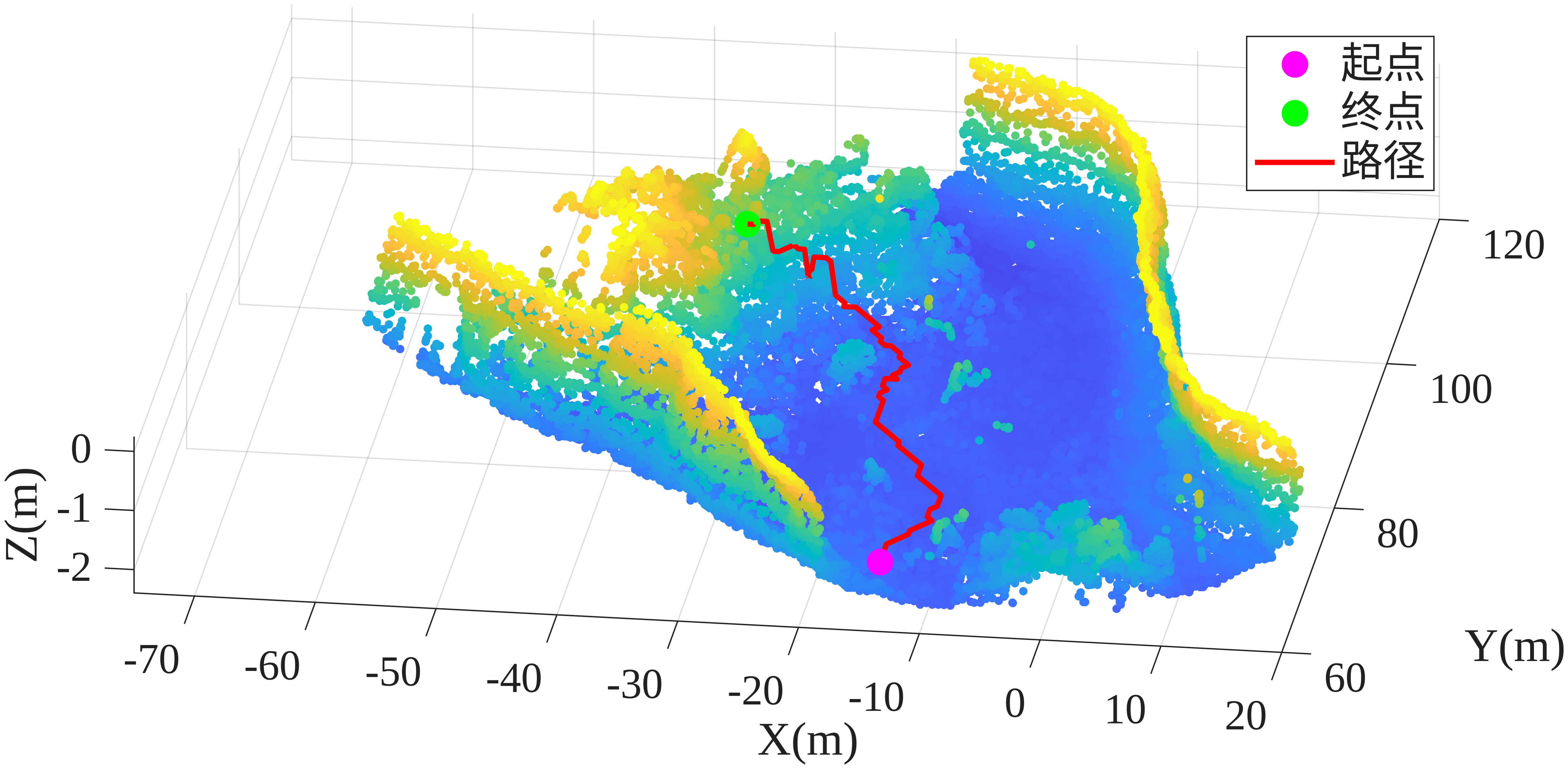}
    \hspace{.2in}
    \caption{Planned path in Scenario 1.}
    \label{fig25}
\end{figure}

\begin{figure} [H]
    \centering
    \includegraphics[height=2.40in]{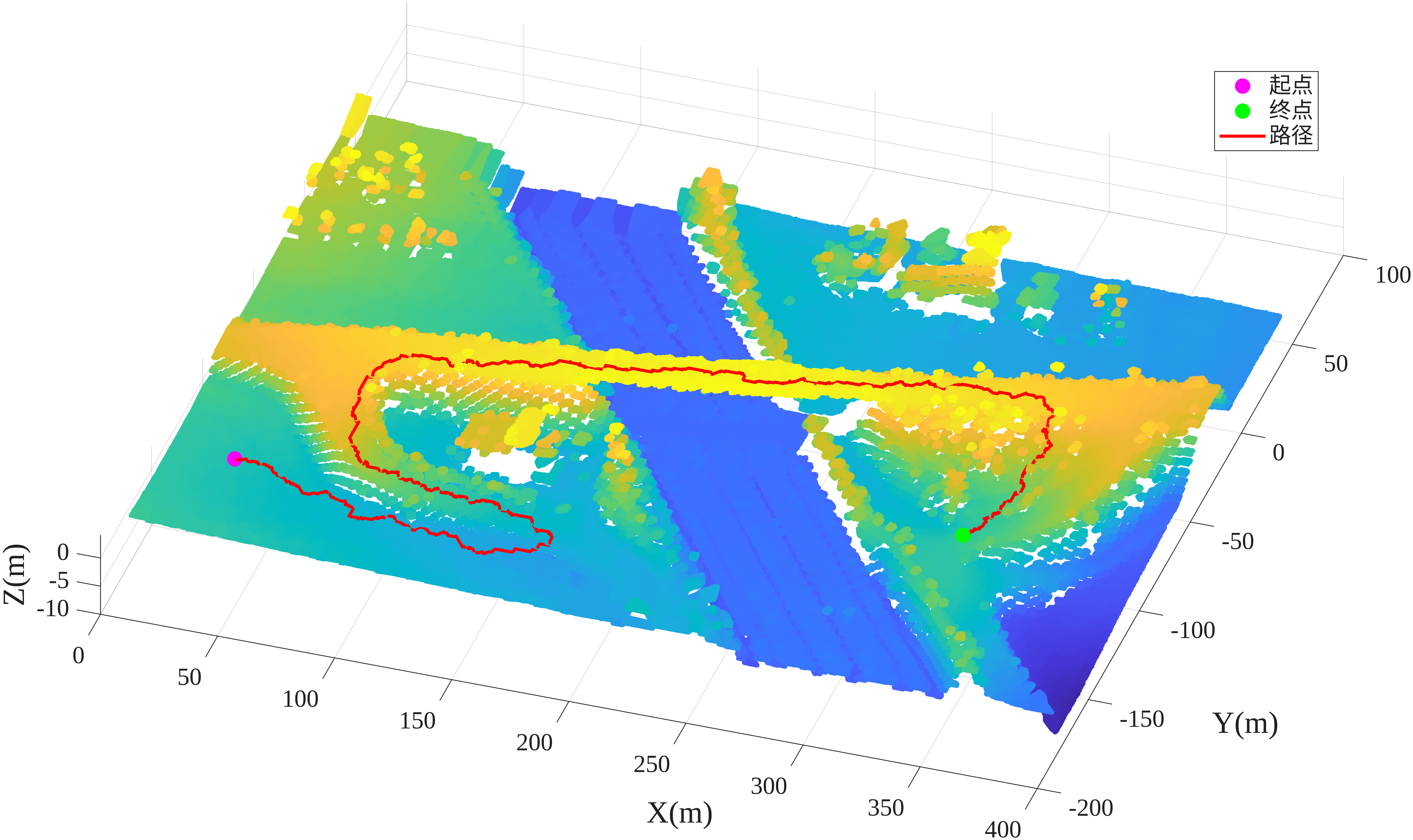}
    \hspace{.2in}
    \caption{Planned path in Scenario 2.}
    \label{fig26}
\end{figure}

\begin{figure} [H]
    \centering
    \includegraphics[height=2.40in]{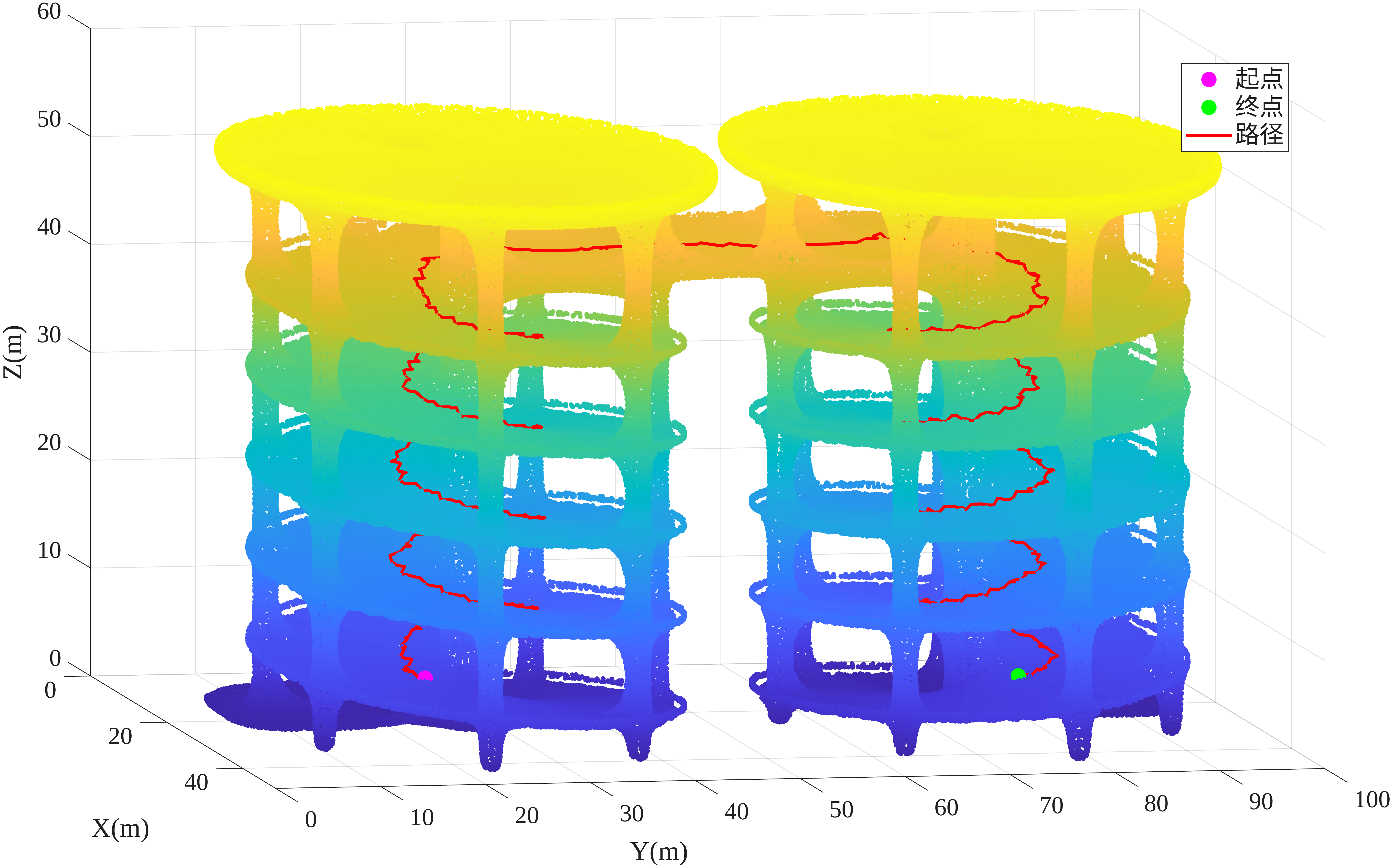}
    \hspace{.2in}
    \caption{Planned path in Scenario 3.}
    \label{fig27}
\end{figure}

Three pairs of start points and endpoints are set in the experimental maps separately. The planned trajectories are presented in Figure \ref{fig25}-\ref{fig27}. Figure \ref{fig25} shows the path in the unstructured, highly irregular terrain. The endpoint was intentionally placed on the top of a small hill. The path keeps away from all extruded rocks on the way and chooses a short path to climb up. Figure \ref{fig26} shows the planning performance in the map with a bridge. The start point and endpoint are selected to be on the two sides of the river separately. The proposed planning algorithm is able to cope with the cluttered environment and to find the only feasible way, the bridge, from one side to the other. Figure \ref{fig27} presents the planned trajectory from the entrance to the exit of a parking garage. Albeit the complex geometric shapes in the building, the path climbs up and travels down over several floors to connect the start point and endpoint, which is the most reasonable and feasible driving path in the garage. The adaptivity of the proposed multi-level 3D path planning method is proven.

Additionally, the proposed method enables explicit control over the trade-off between 
path length and traversability through the hyperparameter $\lambda$ in the travel 
cost function (Eq.~\ref{eq13}). A larger $\lambda$ assigns greater weight to 
traversability, resulting in paths that favor smoother, less steep terrain at the 
expense of potentially longer distances. Since the total travel cost is computed as 
the sum of individual node costs along the path, the appropriate value of $\lambda$ 
depends on the map resolution and voxel size. Therefore, $\lambda$ should be tuned 
for each application based on the specific terrain characteristics and robot 
constraints.

\begin{figure} [H]
    \centering
    \includegraphics[height=2.0in]{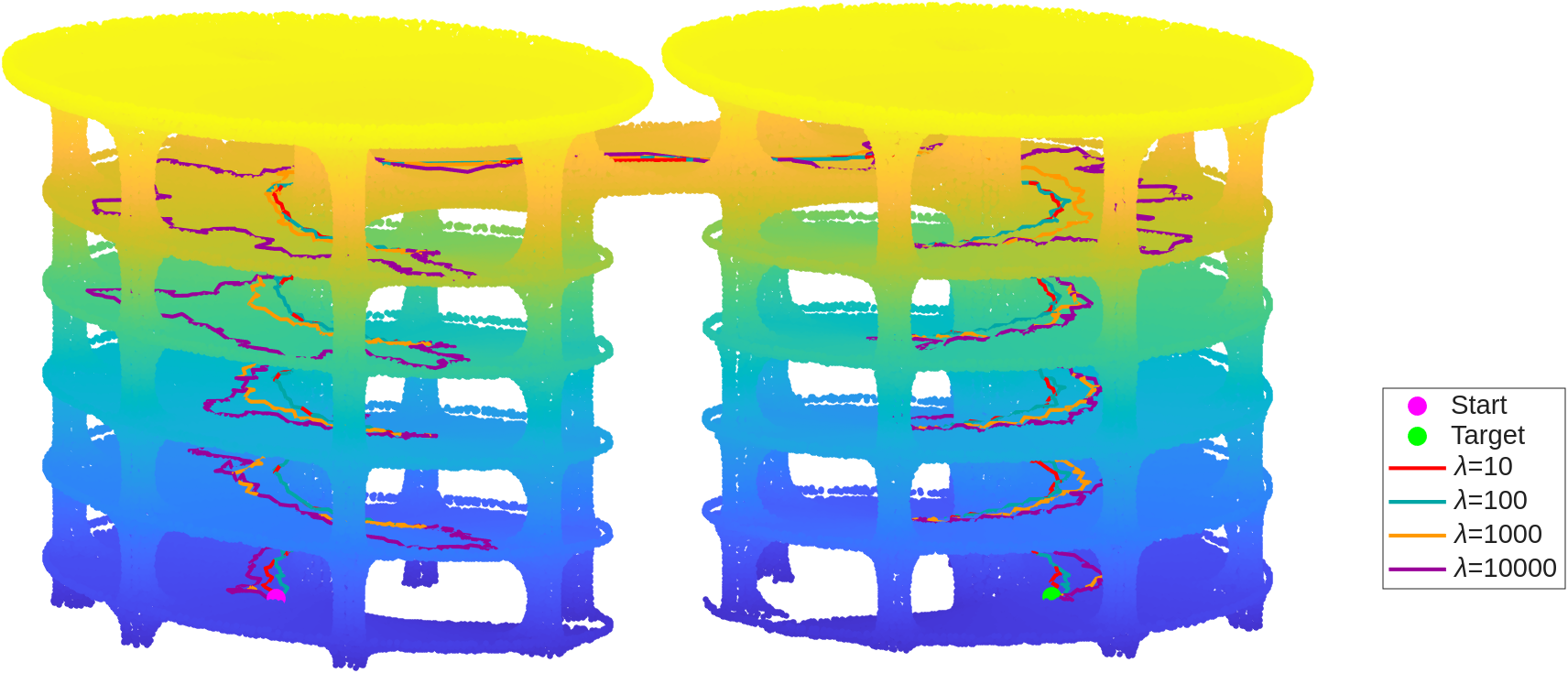}
    \hspace{.2in}
    \caption{Sensitivity analysis of hyperparameter $\lambda$ in Scenario 3 (parking 
    garage). Paths are color-coded by $\lambda$ value: red ($\lambda = 10$), cyan 
    ($\lambda = 100$), orange ($\lambda = 1000$), and purple ($\lambda = 10000$). Higher $\lambda$ 
    values produce longer paths that avoid steep inclines, demonstrating the trade-off 
    between path length and traversability. Path statistics: $\lambda = 10$: 
    $L = 318.3$ m; $\lambda = 100$: $L = 348.3$ m, $\lambda = 1000$: $L = 531.2$ m, $\lambda = 10000$: $L = 954.5$ m.}
    \label{lambda_sensitivity_analysis}
\end{figure}

To demonstrate this trade-off, we conduct a sensitivity analysis by varying $\lambda$ 
in Scenario 3 (the multi-story parking garage). Results are shown in 
Figure~\ref{lambda_sensitivity_analysis}. As $\lambda$ increases, the planner 
generates progressively longer paths that avoid steep slopes and prioritize 
low-traversability-cost regions. For instance, at $\lambda = 10$, the path takes 
the most direct route with length $L = 318.3$ m but traverses moderately steep ramps. 
At $\lambda = 10000$, the path extends to $L = 954.5$ m but maintains 
significantly lower average slope, demonstrating 
the method's ability to adapt planning behavior to user-defined safety preferences.

\begin{figure}[h]
    \centering
    \includegraphics[height=2.00in]{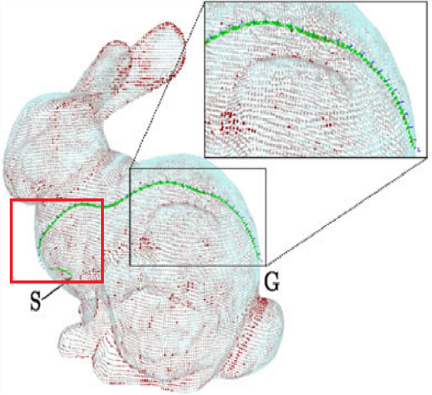}
    \hspace{.2in}
    \caption{The path planning result in \cite{krusi2017driving}. Light blue points denote low roughness, gray points denote high roughness, and dark red ones denote untraversable. }
    \label{fig29}
\end{figure}

\begin{figure}[h]
    \centering
    \includegraphics[height=2.00in]{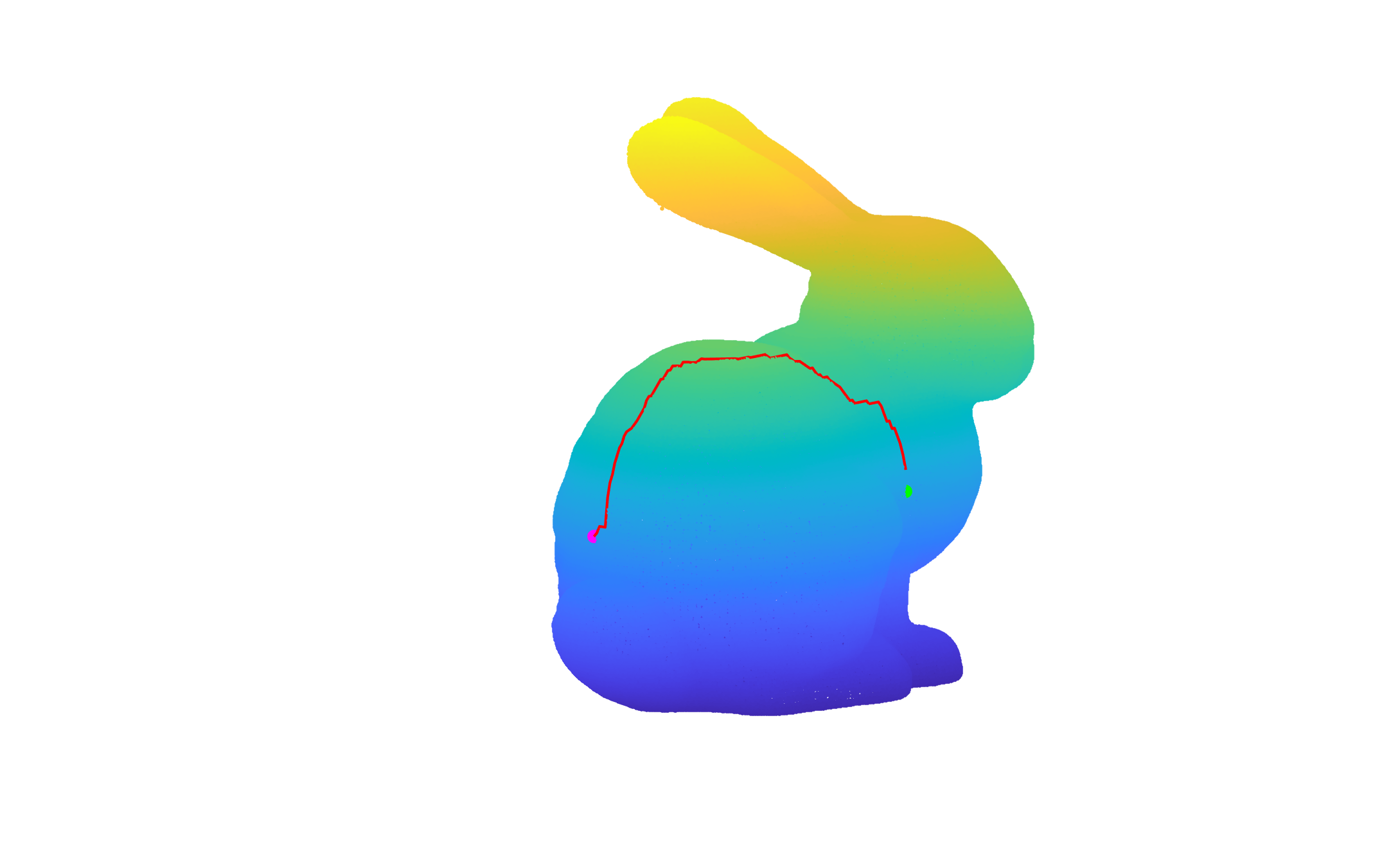}
    \hspace{.2in}
    \caption{The path planning result using our method with similar choice of starting and target points. }
    \label{fig30}
\end{figure}

\begin{figure}[h]
    \centering
    \includegraphics[height=2.00in]{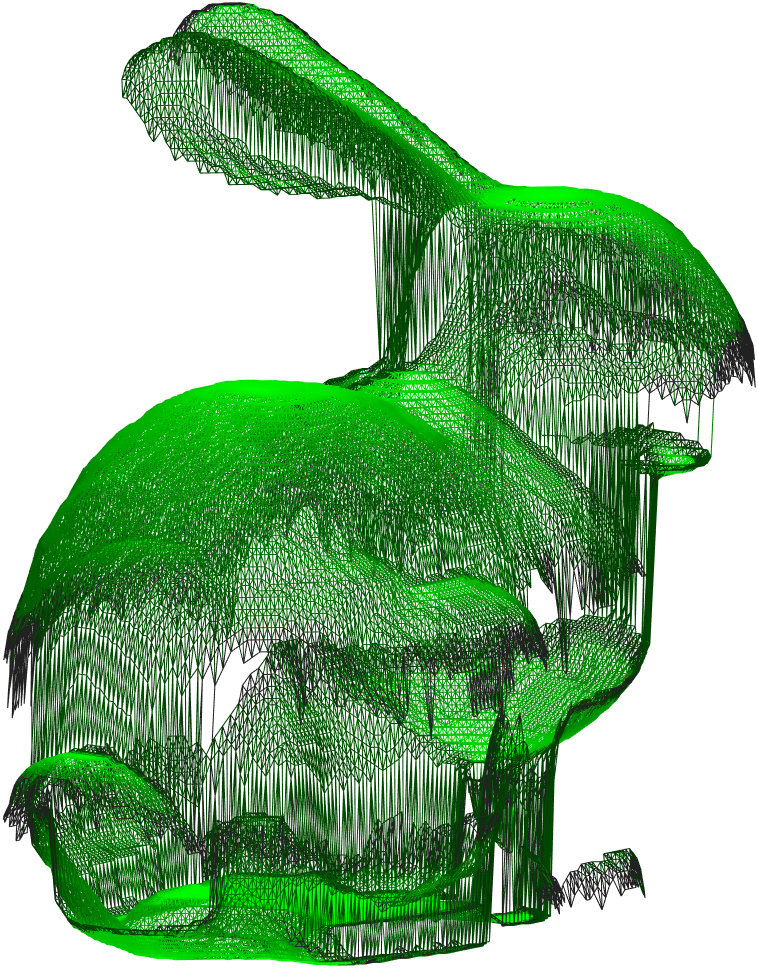}
    \hspace{.2in}
    \caption{The traversability map of the bunny.}
    \label{fig31}
\end{figure}
\subsection{Results comparison with previous work}

To demonstrate the advantages of the proposed method, we conduct a comparative evaluations against recent state-of-the-art approaches for multi-level terrain 
planning. We first present a qualitative comparison with a 3D Sampling-Based planner  
\cite{krusi2017driving} on the Stanford Bunny dataset, then provide quantitative 
comparisons with two recent multi-layer planners on scenarios from 
Section~\ref{sec:case_study_online_database}.

\subsubsection{Comparison with 3D Sampling-Based Planning}

Krüsi et al. \cite{krusi2017driving} perform path planning directly in full 3D space 
by fitting robot-sized planar patches to point clouds and evaluating the local distribution of the map points, which is similar to our environment-robot interaction analysis method. Inherently, their work is path planning in full 3D space, which can be time-consuming. Therefore, they used two-stage path planning to solve the long-distance problem. In our article, since the WTG is a maintained network with discrete voxels on 2D surfaces, the requirement for storage and calculation is much less. Therefore, the path planning proceeds in one stage to seek a globally optimal solution.

The result for the path planning in \cite{krusi2017driving} is shown in Figure \ref{fig29}. 
The robot dimensions were set to $1.3\times 0.7 \times 1.2$ m, with a maximum allowable roll angle of $0.18$ rad and pitch angle limits of $[-0.3, 0.35]$ rad. The maximum admissible terrain roughness (step height) was limited to $0.08$ m, and the maximum trajectory curvature was constrained to $2 ~\text{m}^{-1}$.
Since the surface of the bunny has very steep angles and over-hanging conditions, they relaxed the allowable tilt angle of the mobile vehicles. With similar choices of the starting and ending points as well as relaxation on tilt angle, our path planning method gives a result as in Figure \ref{fig30}. The details inside the black block of Figure \ref{fig29} show that the path on the back of the bunny is reasonable by Krusi’s method. It takes a short way to climb up and avoid hazards. However, due to the two-stage procedure, the path in the red block is clearly unnecessarily detoured, compared to our result in Figure \ref{fig30}. This segment of the path is not only longer in Figure \ref{fig29} but also contains face-up situations. In this aspect, our method has merits in higher calculation efficiency and shorter path length for complex situations.

However, it does not mean these merits are with no cost. The WTG of the bunny is shown in Figure \ref{fig31}. Since our method needs to slice the surfaces vertically into multiple levels, it inevitably causes disconnection where the bunny surface has overhanging structures. Therefore, our method cannot find a way from the back of the bunny to its front chest. Though the surface of the bunny is continuous, its geometric relationship is cut apart by the multi-level structure of the WTG. This defect, however, is not problematic for most UGVs, since they cannot travel on over-hanging surfaces upside down.

\subsubsection{Quantitative Comparison with Recent Multi-Layer Planners}

To rigorously evaluate the proposed method against state-of-the-art approaches for 
multi-level terrain planning, we conduct comparative experiments with two recent 
methods: (1) Surfel-Based State-Space Planner (SBSP) by Atas et al. 
\cite{atas2022elevation}, which constructs a 3D surfel representation through RANSAC 
plane fitting and employs sampling-based planning with constraint-aware sampling 
strategies, and (2) Occupancy Map with Downward Projection (OMDP) planner by Wang 
et al. \cite{wang2023towards}, which uses a 3D voxel grid with a Valid Ground Filter 
and downward projection strategy to map robot poses to traversable surfaces. These 
methods represent distinct paradigms in multi-level planning: true 3D volumetric 
approaches versus 2.5D heightmap-based methods. Our multi-level 2.5D representation 
aims to bridge this gap, achieving the representational completeness of 3D methods 
while maintaining computational efficiency closer to 2.5D approaches.

The comparison is performed on Scenarios 2 and 3 from Section~\ref{sec:case_study_online_database}, 
which contain representative multi-level structures: a highway bridge with vertical 
occlusion (Scenario 2) and a multi-story parking garage with stacked floors connected 
by spiral ramps (Scenario 3). These scenarios challenge different aspects of multi-level 
planning—handling overlapping surfaces at identical horizontal coordinates and managing 
inter-level transitions through continuous ramps. All methods operate on identical raw 
point cloud data with the parameter settings specified in Table~\ref{table1}. 

\begin{figure}[t]
\centering
     \begin{subfigure}[t]{0.45\textwidth}
         \centering
         \includegraphics[width=\textwidth]{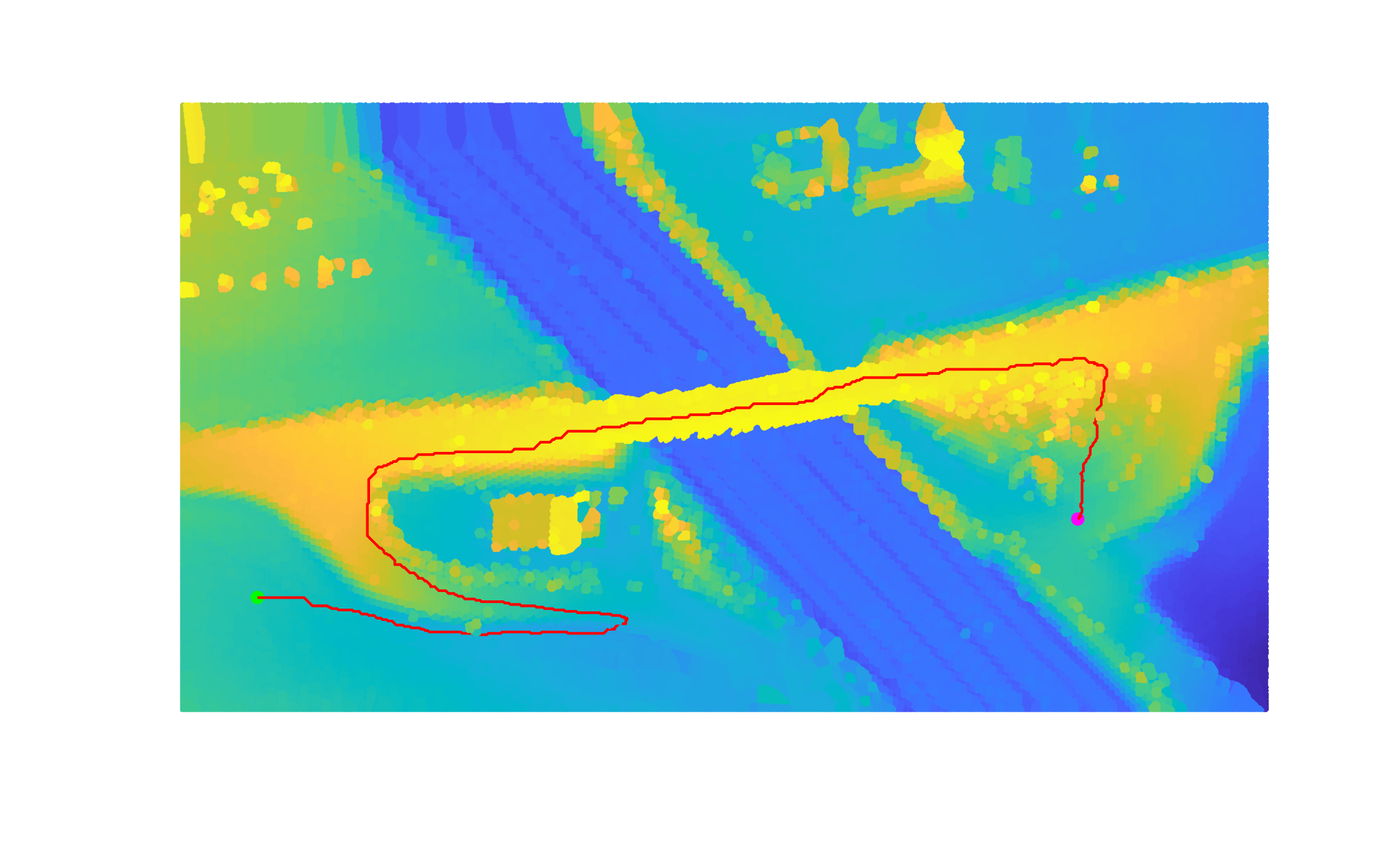}
         \caption{}
         \label{fig:bs_bridge_1}
     \end{subfigure}
     \begin{subfigure}[t]{0.45\textwidth}
         \centering
         \includegraphics[width=\textwidth]{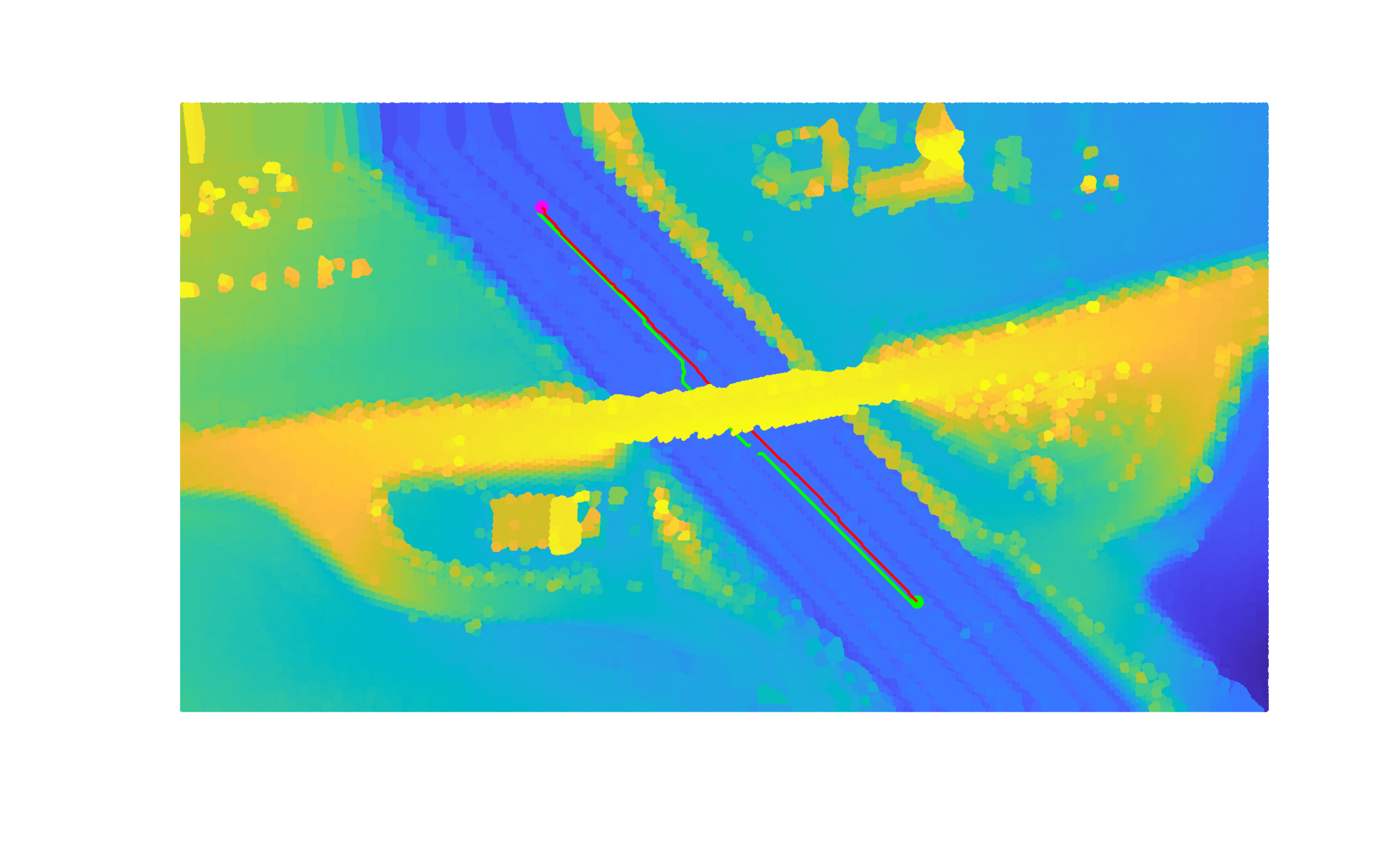}
         \caption{}
         \label{fig:bs_bridge_2}
     \end{subfigure}
     \\
     \begin{subfigure}[t]{0.6\textwidth}
         \centering
         \includegraphics[width=\textwidth]{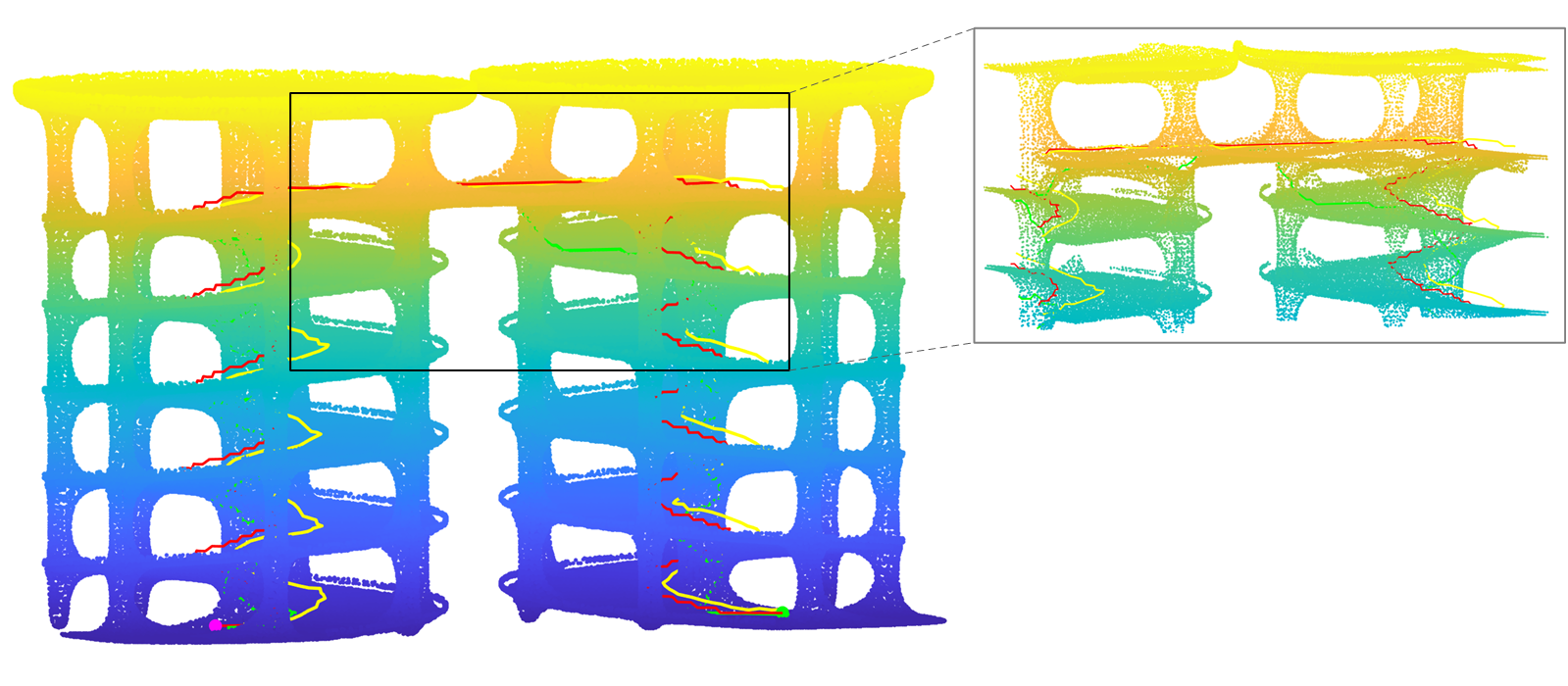}
         \caption{}
         \label{fig:bs_parking}
     \end{subfigure}
     \caption{Baseline comparison results on Scenarios 2 and 3. The proposed method 
demonstrates robust performance across multi-level structures, while competing 
approaches exhibit fundamental limitations: OMDP cannot handle vertical occlusion, and 
SBSP requires severe data reduction to achieve computational 
tractability, sacrificing path quality. The red, green, and yellow paths correspond to the trajectories 
generated by the proposed method, OMDP, and SBSP, respectively. (a) Scenario 2, example 1. (b) Scenario 2, example 2. (c) Scenario 3.}     \label{fig:baseline_comparison}
\end{figure}

For Scenario 2, two test cases evaluate different aspects of bridge navigation. In Case 
2a, start and goal positions are placed on opposite sides of the bridge, requiring the 
path to traverse over the bridge deck. In Case 2b, both positions are located beneath 
the bridge structure, testing the ability to plan along the lower road level while 
correctly recognizing the bridge geometry above. The proposed method successfully finds 
valid paths for both cases, as shown in Figure~\ref{fig:baseline_comparison}(a) and 
(b). In Case 2a, the planner correctly routes the vehicle onto 
the bridge approach, across the deck, and down to the goal on the opposite side, with a 
total path length of 634.9 m computed in 43.6 s. In Case 2b, the path follows the road 
beneath the bridge, maintaining safe clearance from bridge supports, with a path length 
of 203.2 m and planning time of 2.55 s.

In contrast, OMDP planner fails to find a valid path in Case 2a but succeeds in 
Case 2b. This asymmetric performance stems from the fundamental limitation of the 
downward projection strategy employed by Wang et al. \cite{wang2023towards}. The method 
projects each spatial point vertically downward to the nearest surface below and 
evaluates terrain properties at that projected location. This approach implicitly 
assumes a single-valued height function $z = f(x,y)$, where each horizontal coordinate 
maps to at most one traversable surface. Consequently, when the robot is positioned on 
the ground beneath a bridge, downward projection correctly identifies the road surface; 
however, when planning requires transitioning from ground level to the bridge deck at 
the same horizontal location, the representation cannot simultaneously encode both 
surfaces. The planner fails in Case 2a because it cannot recognize the bridge deck as a 
distinct, accessible level above the road. Case 2b succeeds with a path length of 202.7 m and
planning time of 1.6 s because both start and goal reside on the same lower surface, 
avoiding the need to represent vertical occlusion.

The SBSP planner incurs severe computational overhead when applied to maps covering large spatial areas (e.g., 788,054 points over a $400×220 m^2$ area in Scenario 2). The surfel construction process requires RANSAC plane fitting for each sampled point and k-nearest neighbor graph construction across all surfels, resulting in $O(N^2 \log N)$ complexity. Map construction did not complete within a 10-minute timeout, leading to failure in both test cases for Scenario 2.

Scenario 3 evaluates the path planning ability through a multi-story parking garage, where start and 
goal positions are placed at the entrance and exit of the parking building. The optimal 
path must identify and follow the spiral ramp connecting different levels. Our method 
successfully plans a path of length 348.3 m in 10.4 s, correctly 
following the spiral ramp to transition between floors and navigating across the 
buildings to reach the exit. The SBSP planner also succeeds, finding a valid path of 
386.3 m in 7.1 s. However, the OMDP planner exhibits 
catastrophic failure in this scenario. The downward projection strategy introduces ambiguity 
in regions where the ceiling of the lower floor and the floor of the upper level are vertically 
adjacent. As a result, these surfaces are incorrectly merged, producing a path that erroneously 
traverses the ceiling of the lower floor instead of ascending the ramp to the upper level, as 
illustrated in Figure~\ref{fig:baseline_comparison}(c). This failure highlights the fundamental unsuitability 
of 2.5D single-valued representations for true multi-story environments with vertically stacked 
navigable surfaces.

Beyond planning performance, the efficiency of the underlying map representations is 
evaluated in Table~\ref{qualitative_compare_table2}. We measure (1) surface retrieval 
time, defined as the time required to access all traversable surface voxels or surfels 
from the stored representation, which is critical for rapid replanning in dynamic 
environments, and (2) memory footprint for storing the complete map structure. The 
proposed ML-SkiMap structure achieves surface retrieval in 0.032 s for Scenario 2, 
representing a 2.2$\times$ speedup over OctoMap (0.070 s) and 3.2$\times$ speedup over 
the surfel-based representation (0.103 s). This performance advantage stems from the 
SkipList data structure's $O(\log n)$ access time and the explicit encoding of surface 
layers, which eliminates the need to traverse the full 3D volume (as in OctoMap) or 
query a dense k-NN graph (as in surfel-based methods).

Memory efficiency is similarly improved through the variable-resolution encoding 
described in Section~2.2. For Scenario 2, the ML-SkiMap requires 12.6 MB of storage 
compared to 43.4 MB for OctoMap and 51.3 MB for the surfel representation, representing 
reductions of 3.4$\times$ and 4.1$\times$, respectively. This compression is achieved 
through curvature-based adaptive downsampling (Eq.~\ref{eq4}), which retains high point 
density in geometrically complex regions while aggressively reducing 
points in flat areas. Importantly, despite this substantial data 
reduction, the planning success rate remains 100\%, indicating that the preserved 
geometric information is sufficient for accurate traversability assessment and path 
generation.

\begin{table}[]
\centering
\begin{tabular}{|llll|}
\hline
\multicolumn{1}{|l|}{}     & \multicolumn{1}{l|}{SBSP-planner \cite{atas2022elevation}} & \multicolumn{1}{l|}{OMDP-planner \cite{wang2023towards}} & Ours   \\ \hline
\multicolumn{4}{|c|}{Scenario 2 over bridge}                                                                                \\ \hline
\multicolumn{1}{|l|}{Path length (m)}      & \multicolumn{1}{l|}{/}             & \multicolumn{1}{l|}{/}             & 634.9 \\ \hline
\multicolumn{1}{|l|}{Computation time (s)} & \multicolumn{1}{l|}{/}             & \multicolumn{1}{l|}{/}             & 43.6   \\ \hline
\multicolumn{4}{|c|}{Scenario 2 under bridge}                                                                               \\ \hline
\multicolumn{1}{|l|}{Path length (m)}      & \multicolumn{1}{l|}{/}             & \multicolumn{1}{l|}{202.7}             & 203.2   \\ \hline
\multicolumn{1}{|l|}{Computation time (s)} & \multicolumn{1}{l|}{/}             & \multicolumn{1}{l|}{1.6}             & 2.55   \\ \hline
\multicolumn{4}{|c|}{Scenario 3}                                                                                            \\ \hline
\multicolumn{1}{|l|}{Path length (m)}      & \multicolumn{1}{l|}{386.3}             & \multicolumn{1}{l|}{/}             & 348.3  \\ \hline
\multicolumn{1}{|l|}{Computation time (s)} & \multicolumn{1}{l|}{7.1}             & \multicolumn{1}{l|}{/}             & 10.4   \\ \hline
\end{tabular}
\label{qualitative_compare_table1}
\caption{Path planning performance comparison across test scenarios. Planning time 
includes both map query and graph search. ``/'' indicates planning failure (no valid 
path found within timeout or geometrically invalid result). }
\end{table}

\begin{table}[]
\centering
\begin{tabular}{|l|l|l|l|l|}
\hline
                  & \multicolumn{1}{c|}{\begin{tabular}[c]{@{}c@{}}Surface Retreival time (s),\\ Scenario 2 \end{tabular}} 
                  & \multicolumn{1}{c|}{\begin{tabular}[c]{@{}c@{}}Map size (MB),\\ Scenario 2\end{tabular}} 
                  & \multicolumn{1}{c|}{\begin{tabular}[c]{@{}c@{}}Map size (MB),\\ Scenario 3\end{tabular}} \\ \hline
Surfel-based Map \cite{atas2022elevation} &   0.103      &     51.3       &  9.4        \\ \hline
OctoMap \cite{wang2023towards}            &   0.070      &     43.4      &   8.1        \\ \hline
ML-SkiMap (Ours)                          &   0.032      &     12.6      &    8.6        \\ \hline                                                                    
\end{tabular}
\caption{Comparison of map retrieval efficiency and memory usage. 
Surface retrieval time denotes the time required to access all traversable surface 
elements from the stored map representation. Map size indicates the total memory 
required to represent the environment, including the original point cloud data and 
the associated voxel- or surfel-based map structures used for planning.}
\label{qualitative_compare_table2}
\end{table}

\section{Conclusion}
This article presents a practical approach using point cloud data to quickly and safely plan paths for UGVs working in complex environments. It is one of the few research articles with path planning directly on point cloud map, and it is the first one of this kind to specifically solve the multi-level problem. 

We propose a combination of the efficient multi-level SkiMap point cloud management method and variable resolution data simplification, which achieves adequate data slim and fast data search. This essentially decreases the storage demand and computational load, which are the two fundamental problems for point-based methods. With the structured voxels in the multi-level SkiMap, a WTG is proposed to present a geometric network among surface levels. The local traversability index is assigned as the weights on the edges of the graph so that the WTG contains information on driving safety and surface connection conditions. With a modified A* algorithm, the globally optimal path connecting any feasible start point and endpoint can be obtained.

The algorithms are demonstrated with both an experimental system and online databases. The experimental tests based on point cloud maps by a mobile robot include indoor environment, motte, and grove. The database tests include the valley, bridge, and parking lot. The results show that the algorithm has high adaptation and effectiveness in various situations. We also compared our method with the state of art work on point-based 3D path planning \cite{krusi2017driving}. Our approach has advantages in fast computation and global optimality, while theirs can handle upside-down driving situations. For general mobile robots or vehicles that can only traverse upon surfaces with limited tilt angles, our method has the full capability to handle complex terrain conditions, including cliffy, overhanging shapes, and multi-level structures.     

We also would like to point out that the outliers in the raw measurement data could potentially corrupt the results. The algorithm to eliminate outliers is not in the scope of this article thus not emphasized, but it needs special care in real applications. Or it could lead to local traversability analysis failures if not well treated.

\section*{Acknowledgements}
This research is financially supported by Jiangxi Research Institute, Beihang University, and the Open Fund of Intelligent Control Laboratory under Grand No. 2024-ZKSYS-KF02-02. The authors gratefully acknowledge Hang Zhao for his valuable support in the execution of the real-world experiments.

\section*{Conflict of Interest}
The authors declare no conflict of interest.

\section*{Data Availability Statement}
The data used in this research are available upon request.

\section*{Author Contributions}
Yangmin Xie, Hang Shi, Yusheng Yang and Yujie Tang designed the research. Yujie Tang, Li Quan and Hao Geng processed the corresponding data. Yangmin Xie, Yujie Tang and Yusheng Yang wrote the first draft of the manuscript. Hang Shi helped to organize the manuscript. Yujie Tang revised and edited the final version.

\nolinenumbers

 \bibliographystyle{plos2015} 
 \bibliography{cas-refs}

%
%
%





\end{document}